\documentclass[letterpaper, 10 pt, conference]{ieeeconf}  % Comment this line out if you need a4paper

\IEEEoverridecommandlockouts                              % This command is only needed if 
\usepackage{amsmath}
\usepackage{amssymb}  
\usepackage{booktabs}
\usepackage{float} % for pdf, bitmapped graphics files
\usepackage{epsfig} % for postscript graphics files
\usepackage{algpseudocode}

\usepackage{siunitx} 
\usepackage[dvipsnames]{xcolor}
\usepackage{algorithm}
\usepackage{hyperref}
\hypersetup{
    colorlinks=true,
    linkcolor=blue,
    urlcolor=blue,
    citecolor=blue,
    pdftitle={SLAMFuse},
    pdfpagemode=FullScreen,
    }
\usepackage{nicematrix}
\usepackage{colortbl}
\newcolumntype{C}{>{\centering\arraybackslash}X}
\usepackage{gensymb}
\usepackage{multirow, array}
\usepackage{caption}
\usepackage{subcaption}
\title{\LARGE \bf FLIP: Flowability-Informed Powder Weighing }

\author{Nikola Radulov$^{1}$, Alex Wright$^{1}$, Thomas Little$^{1}$, Andrew I. Cooper$^{2}$, Gabriella Pizzuto$^{1, 2}$\\
\thanks{$^{1}$ Department of Computer Science, University of Liverpool, UK.}%
\thanks{$^{2}$ Department of Chemistry, University of Liverpool, UK.}} %

\begin{document}
\maketitle
% \vspace{-1em}
\begin{figure*}[t!]
    \centering
    \includegraphics[width=0.7\textwidth]{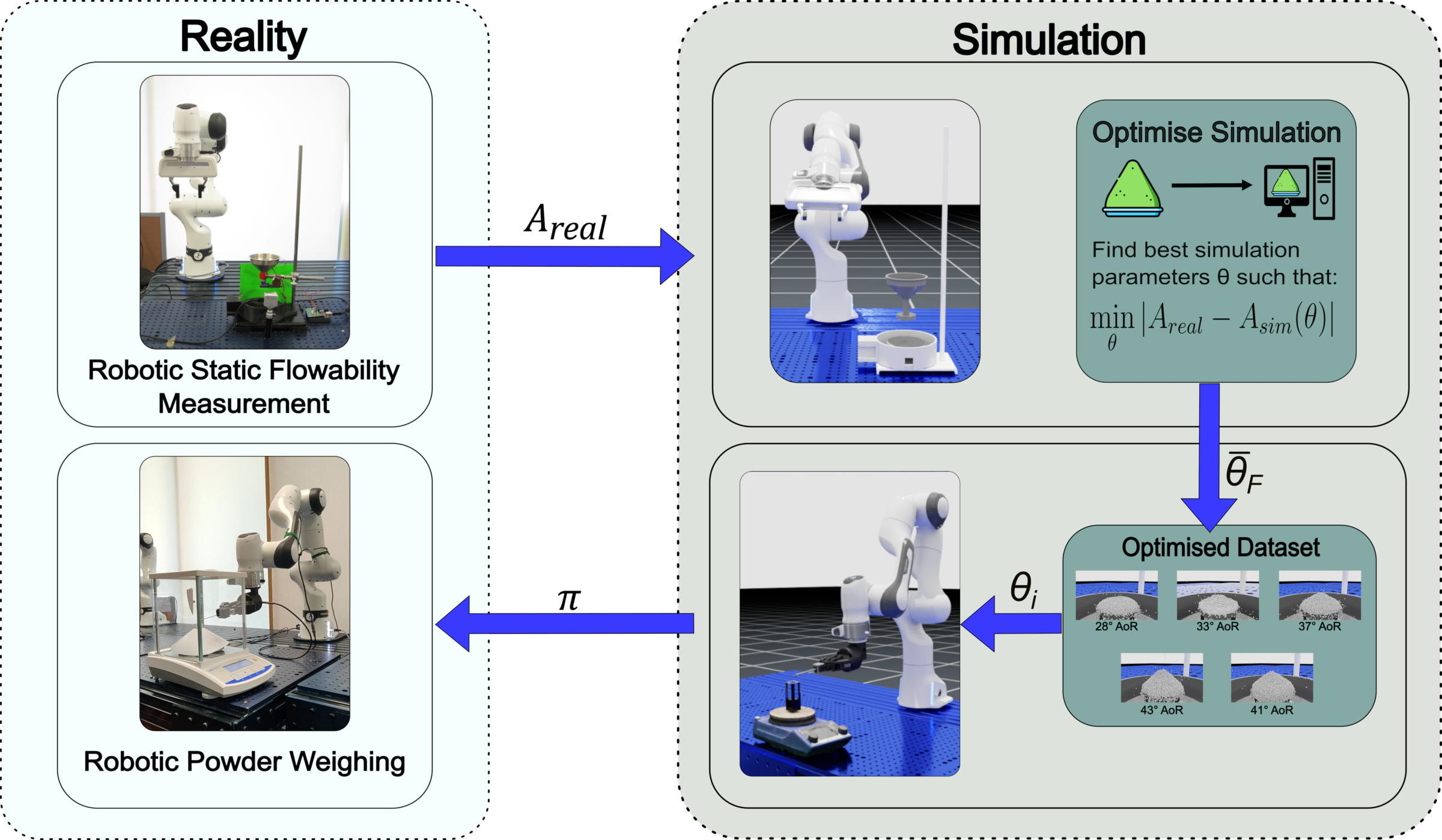}
    % \caption{FLIP Overview: Using Bayesian Optimisations, we learn simulation parameters that allow us to match real-world granular behaviour when measuring the static flowability, of a set of real-world powders, covering a specified flowability range. Using these parameters sets we construct a flowability dataset which we use to train a control agent for powder weighing.  }
        \caption{The \textbf{\textit{FLIP}} framework leverages BO to calibrate simulation parameters using real-world powder flowability data. This flowability-informed powder simulation then serves as a high-fidelity environment for training the robot control policy for powder weighing.}
    \label{fig:flip_pipeline}
\end{figure*}
% \begin{figure*}[h!]
%     \centering
%     \includegraphics[width=0.75\textwidth]{example-image-a}
%     \caption{}
%     \label{fig:overall_block_diagram} 
% \end{figure*}

\begin{abstract}

Autonomous manipulation of powders remains a significant challenge for robotic automation in scientific laboratories.
The inherent variability and complex physical interactions of powders in flow, coupled with variability in laboratory conditions necessitates adaptive automation.
This work introduces \textbf{\textit{FLIP}}, a flowability-informed powder weighing framework designed to enhance robotic policy learning for granular material handling.
Our key contribution lies in using material flowability, quantified by the angle of repose, to optimise physics-based simulations through Bayesian inference.
This yields material-specific simulation environments capable of generating accurate training data, which reflects diverse powder behaviours, for training `robot chemists'.
Building on this, \textit{\textbf{FLIP}} integrates quantified flowability into a curriculum learning strategy, fostering efficient acquisition of robust robotic policies by gradually introducing more challenging, less flowable powders.
We validate the efficacy of our method on a robotic powder weighing task under real-world laboratory conditions.
Experimental results show that \textbf{\textit{FLIP}} with a curriculum strategy achieves a low dispensing error of $2.12 \pm \SI{1.53}\mg$, outperforming methods that do not leverage flowability data, such as domain randomisation ($6.11 \pm \SI{3.92}\mg$).
These results demonstrate \textbf{\textit{FLIP}}'s improved ability to generalise to previously unseen, more cohesive powders and to new target masses.

\end{abstract}
\vspace{-0.5em}

\section{INTRODUCTION} 

%     \item Explain the importance of material discovery: need of new materials to improve on renewable energy production, energy storage, drug development, finding new environmental friendly materials, and making goods overall cheaper through better manufacturing and materials
% Accelerating chemical and material discovery is critical to future societal and industrial impact. 
% We urgently require the ability to design and discover the materials needed to build a more sustainable, prosperous and healthy future. 
% Stemming from an increasing demand of challenges arising from global warming and the COVID-19 pandemic, AI-driven robotics is essential to accelerate scientific discovery. 
% The future of chemical discovery, \textit{e.g.,} using generative AI methods that can identify novel molecules and materials classes with tailored compositions, structures, and functions, will heavily rely on the data generated from autonomous robotic experiments~\cite{Abolhasani2023}.

% \item Highlight the benefit of robotic scientists: they carry out tedious tasks,giving the human scientist freedom to concentrate on mentally stimulating ones
% \item Laboratory robots have been used primarily for sample transportation \cite{fakhruldeen2022archemist}\cite{Burger2020AMR} as highly dexterous tasks are still quite challenging to automate.
Robotic systems are increasingly used in laboratories to relieve human researchers of tedious and repetitive tasks, allowing them to focus on higher-level scientific reasoning. 
One embodiment of this is mobile autonomous robotic chemists, which have achieved success in different chemistry domains~\cite{Tom2024}, such as photocatalysis~\cite{Burger2020AMR}, solid-state powder x-ray diffraction~\cite{Lunt2024}, and organic synthesis~\cite{Dai2024}.
While effective and reliable for sample transportation, robotic methods have yet to demonstrate robust and adaptive scalability for more complex tasks in this context.
Consequently, more dexterous lab tasks still require the skill and specialised knowledge of human researchers.
% Robotic solutions in this area are still in their infancy, with very few highly specialised and hand-engineered solutions.
% This limits potential applications especially in the realm of early-stage materials discovery where sample properties are heterogeneous and unpredictable.
% \item  Powders play a crucial role within material discovery, being essential in techniques like additive manufacturing, catalysis and battery development. 
% \item powder manipulation is challenging from the perspective of robotic manipulation, due to their complex dynamics, making it difficult to predict their behavior
% \item the task of powder weighing  exemplifies these challenges: here the robot needs to be able to adapt to a plethora of powders with varying physical properties, from highly cohesive to flowable powders
% \item Current powder weighing systems tend to suffer from generalisation issues, struggling to adapt to this high variety of physical properties 
% \item Mention how cohesive , compressible and hygroscopic powders cause issues , like blockage in \cite{jiang}

Developing robotic chemists that are capable of handling different materials with a wide range of properties, such as granular and powdery materials, would allow us to bring autonomous materials discovery to a new level. For example, robots could carry out end-to-end chemical synthesis, from sample preparation and reaction execution to transfer of samples for characterisation.
Despite their central role in both materials chemistry and organic synthesis, powders remain difficult to manipulate reliably with robotic systems. 
This difficulty stems from the complex, non-linear dynamics of powders that are highly sensitive to material properties in real-world laboratory conditions.   For example, many powders are hygroscopic, meaning that their properties can change signficantly under humid conditions.
A fundamental task that underpins chemistry and materials science is powder weighing.
% The inherent challenge within automating this task would depend on the particular properties of a given sample, that could range from highly cohesive and hygroscopic ones that tend to clump or block dispensing mechanism~\cite{jiang}, to dry, free-flowing powders that behave more predictably. 
The central challenge in automating this task is that properties of a given sample can range from highly cohesive and abrasive powders, to dry, free-flowing powders that behave more predictably. 
Existing powder weighing systems~\cite{jiang}, while highly effective in specific domains, often struggle with this level of generalisation, and they are not necessarily designed to accommodate the diversity of material behaviours present across different applications in materials research.
Approaches that have used general purpose robots to address this include using dual-arm manipulators with fuzzy logic control~\cite{jiang}, or employing a single manipulator guided by a deep reinforcement learning algorithm~\cite{powder_weighing_tanaka}.
While the method by Jiang et al.~\cite{jiang} used a rule-based system with binned particle size input, neither of these methods addressed how multiple material properties can be embedded directly within a learning framework to generalise more effectively to entirely new materials.
% \item We aim to leverage material properties,  in our case flowability, to improve the success rate within the task of powder weighing. 
% \item We combine this with a sim-to-real training approach
% \item Similar to \cite{powder_weighing_tanaka} we decouple powder dumping from the rest of the task, and we concentrate our efforts on this
% \item mention ? that for tool manipulation and scooping we  use manually defined controllers
% \item Firstly, construct a process for diminishing the gap between simulated and real-world environments focusing on powder flowability.
% \item We then build an optimised set of simulated training powders used to train reinforcement learning policies for powder dumping
% \item We deploy our policy on a real life environment
% \item We compare this to DR policy 

In this work, we propose a method to enhance the reliability of robotic powder weighing within automated experimental workflows by explicitly leveraging knowledge of material properties, specifically powder flowability. 
This property defines how readily a powder flows under gravity or applied stress~\cite{Prescott2000}, influencing its handling characteristics (\textit{e.g.,} during dispensing and dosing).
% It also underpins key operations such as transportation, mixing, and processing across a wide range of industries \cite{SINGH2020101388, flour_flow}.
% Similar to the approach in \cite{powder_weighing_tanaka}, we decouple powder dumping from the rest of the task and focus our efforts on this key subproblem. 
% For other components such as tool manipulation and scooping, we employ manually defined controllers. 
First, we construct a framework to reduce the simulation-to-reality gap for powder manipulation.
This involves using automated Angle of Repose (AoR) measurements from real powders to optimise key simulation parameters (\textit{e.g.}, friction, cohesion, adhesion).
The resulting flowability-informed simulation environment provides a training dataset representative of various powders, enabling the training of a Reinforcement Learning (RL) policy for precise powder dispensing. 
We explore curriculum learning~\cite{Bengio2009}, increasing material cohesiveness as training progresses, and evaluate our approach through zero-shot transfer performance on real powders compared to a Domain Randomisation (DR) baseline~\cite{powder_weighing_tanaka}.
Finally, we zero-shot transfer this flowability-informed policy to a physical robot and evaluate its performance on real powders.

In summary, the contributions of this work are:
\begin{enumerate}
    \item The design and implementation of \textbf{\textit{FLIP}}, a novel framework integrating material property measurement, flowability-informed granular materials simulation, and policy learning for improving robotic powder weighing in laboratory automation.
    \item A robotic system for accurate AoR measurement, validated against manual methods.
    \item A simulation calibration framework that leverages measured material flowability data to improve granular material dynamics within robotic simulators.
    \item Empirical validation demonstrating an improved robotic policy performance and generalisation for the challenging task of powder weighing, validated through simulation and real-world experiments.
\end{enumerate}
\vspace{-0.5em}
\section{Related Work}
\label{sec:related_work}
% Should have 3 different sections: 
\subsection{Robotics for Laboratory Automation}
% Present recent notable achievements in laboratory automation
\label{ssec:robotics_lab_automation_rel_work}
% Autonomous robotic systems are fundamental towards accelerating laboratory experiments. 
%% robotic chemists for workflows
The rapid progress in robotics for a multitude of everyday tasks has led to increased interest in the adoption of robotic platforms for chemistry laboratory environments, with mobile~\cite{Burger2020AMR,Zhou2025} and fixed-base manipulators~\cite{Darvish2025} spearheading this endeavour.
%% robotic chemists for skills/tasks
While these efforts can automate routine tasks, `robotic chemists' still lack the dexterity and skill of human chemists.
Recent works have focused on introducing sensory feedback for glassware manipulation~\cite{Butterworth2023}, injection automation~\cite{injection} and microplates~\cite{Scamarcio2025}.
%% robotic chemists for material manipulation
Beyond glassware manipulation, human chemists can handle heterogeneous materials with varying physical properties.
% Thus, `robot chemists' should also be capable of this.
Some works have looked at material manipulation in `robot chemists' across different lab tasks, for example, sample scraping~\cite{pizzuto2022accelerating}, grinding~\cite{grinding} and powder weighing~\cite{jiang},~\cite{powder_weighing_tanaka}.
%% gap in literature and how our method addresses this
Here, the authors focused on automating the tasks using either data-driven (model-free, off-policy deep reinforcement learning) or rule-based methods.
While some approaches may have used physical properties as inputs, they did not embedded them directly within the learning framework itself.
Here, we concentrate our research endeavours on adding material properties directly into the learning paradigm towards improving robotic manipulation in laboratory environments. 

%Powder manipulation papers: \cite{grinding} use a combination of visual and audio feedback in order to control powder grinding process. 

% For powder weighing \cite{jiang} uses a dual arm manipulator and a fuzzy logic controller to accurately weight samples as small as 20mg. For even smaller samples, \cite{powder_weighing_tanaka} make use of reinforcement learning and domain randomisation  

% \begin{itemize}
%     % \item Robots doing full chemistry task \cite{Burger2020AMR} experiment to find photocatalists ,  full automated system for polymer press
%     \item other work done usually for more specific tasks, usually part of larger chemistry experiments: injecting-\cite{injection} scraping-\cite{pizzuto2022accelerating} vial manipulation-\cite{vial-manipulation} 
%     \item Powder manipulation papers: \cite{grinding} use a combination of visual and audio feedback in order to control powder grinding process. 
%     For powder weighing \cite{jiang} uses a dual arm manipulator and a fuzzy logic controller to accurately weight samples as small as 20mg. For even smaller samples, \cite{powder_weighing_tanaka} make use of reinforcement learning and domain randomisation  
% \end{itemize}
\subsection{Sim-to-real Gap in Robotic Manipulation of Materials}
\label{ssec:sim2real_physics_related}

Simulators are vital for developing learning-based `robotic chemists' and generating cost-effective training data. 
However, their effectiveness in chemistry lab automation hinges on accurately modelling real-world interactions, such as contact-rich manipulation and particle-based simulation, since the inherent differences between simulated and real-world physics (sim-to-real) pose a major hurdle for robotics.
A common strategy to mitigate this is domain randomisation~\cite{tobin2017domainrandomizationtransferringdeep}, where simulation parameters or visuals are varied to improve policy robustness. 
For instance, Kadokawa et al.~\cite{powder_weighing_tanaka} used a domain-randomised simulator to train an RL agent for powder weighing, achieving sub-milligram accuracy but lacking explicit material modelling for broader generalisation across powders with different properties. 
Sim-to-real transfer methods have also enabled policy deployment in other high-precision tasks such as deformable object manipulation~\cite{deformable_s2r}.
While domain randomisation can improve robustness, achieving accurate simulation of complex physical materials is also critical for reliable skill transfer, particularly for tasks involving sensitive interactions with materials such as powders. 
Recent works calibrate simulation parameters using real-world observations to improve physics fidelity.
Notably, Matl et al.~\cite{Matl2020} inferred granular properties (\textit{e.g.}, friction, restitution) from visual data of formations for tasks like pouring, and Lopez-Guevara et al.~\cite{stir-to-pour} inferred liquid properties through interaction-based probing for pouring optimisation.
Building upon these physics optimisation techniques, and in contrast to prior sim-to-real methods primarily relying on DR, our approach explicitly uses real-world flowability data to inform simulation parameters. 
This enables the creation of material physics-informed environments where knowledge is embedded directly into the training loop for enhanced robotic powder manipulation performance and generalisation.

\section{Methodology}
This work introduces a methodology that improves powder weighing by leveraging material-specific knowledge, specifically powder flowability.
To address the sim-to-real challenge, we develop a framework (Section~\ref{ssec:optimisation_framework}) that optimises simulation parameters (\textit{e.g.,} adhesion, cohesion, friction, etc.) using real-world flowability data, thereby creating realistic training environments for robots to learn how to handle diverse powders.
Using the optimised simulator informed by real-world materials, we introduce \textit{\textbf{FLIP}} (Section~\ref{ssec:FLIP_method}), our framework for powder weighing that leverages a high-fidelity simulator informed by real-world material properties.
% This section outlines our proposed framework for learning powder weighing using material flowability information. 
We define the task of powder weighing as follows: the robot is equipped with a spatula in its end effector, an analytical balance is used for weight feedback, and the target powder is collected in a container (typically using lab glassware).
% We first formulate autonomous robotic powder weighing as a reinforcement learning problem, using data optimised with the method presented in Section~\ref{sssec:optimisation_process}. 

\vspace{-0.5em}
\subsection{Closing the Sim-to-real Gap For Powders Using Flowability Data}
\label{ssec:optimisation_framework}
%%Previous option: REDUCING THE SIM-TO-REAL GAP FOR POWDER FLOWABILITY

%%%%%%%% TEXT TO ADD SOMEWHERE %%%%%%%%%%%%%%%%%%
% Similar to the approach in \cite{powder_weighing_tanaka}, we decouple powder dumping from the rest of the task and focus our efforts on this key subproblem. 
% For other components such as tool manipulation and scooping, we employ manually defined controllers.
%%%%%%%% %%%%%%%% %%%%%%%% %%%%%%%% %%%%%%%% 

% Describe the optimisation process, introduce simulation description +  show tom's real life equivalent

\subsubsection{Problem Formulation}
\label{sssec:optimisation_process}

Let the dynamics of the simulation engine be described by a function $p$ parameterised by $\theta$, which gives the probability distribution over the state $s_t$ at time $t$, given the state $s_{t-1}$ and action $a_{t-1}$ at time $t-1$. 
Let $A$ be a task, or a set of tasks, that can give a quantifiable measure of the physical property to optimise.
Let $A_{real}$ be the measure obtained from executing task(s) $A$ in the real world and $A_{sim}(\theta)$ be the corresponding measure obtained by executing $A$ in the simulation environment.
The reality gap, or error $E$, with respect to the physical property under optimisation is defined as: $E=|A_{real} - A_{sim}(\theta)|$. 
% The dynamics $p$ of our simulation engine directly influence the result of performing $A$ and since $p$ is dependant on $\theta$ we can rewrite our the result of our tasks as being dependant it: $A(\theta)$.
The objective is to find the set of simulation parameters $\theta$ such that $E$ is minimised:
\vspace{-1.2em}

\begin{equation}
    \min_{\theta} |A_{real} - A_{sim}(\theta)|
    \label{eq:sim2realgap}
\end{equation}
\vspace{-1.2em}

\begin{figure}[h!]
    \centering
    \includegraphics[width=0.65\linewidth]{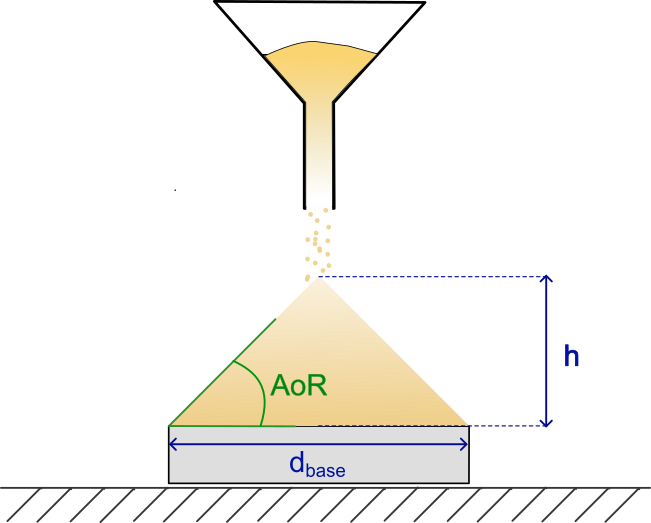}
    \caption{Measurement of static powder flowability using the AoR method. The figure depicts the experimental setup, the formation of the powder pile after flowing through a funnel, and indicates the height measurement used for AoR calculation.}
    \label{fig:static_flowability}
\end{figure}

To optimise the simulation framework with respect to our physical property, powder flowability, the task set $A$ corresponds to the measurement of static flowability~\cite{Liu2024}.
This property is quantified by the AoR, which inversely correlates with static flowability (\textit{i.e.}, a higher AoR results in a lower flowability).
This is measured experimentally using a powder tester following the procedure outlined in the ISO 8398:1989 standard (Fig.~\ref{fig:static_flowability}).
The process involves loading approximately $\SI {50}g$ of powder into a funnel, allowing it to form a pile on a fixed base and using vibration to assist flow through a sieve.
This is followed by measuring the height $h$ of the resulting pile. 
The AoR is then calculated from the measured height and the known base diameter ($d_{base}$) using Equation~\ref{eq:aor}.
% \begin{enumerate}
%     \item Load around 50g of powder into a funnel
%     \item Vibrate the sieve to assist the powders going through 
%     \item Measure the height after all the powder goes out the funnel
%     \item Compute the AOR using the following equation: 
% \end{enumerate}
\begin{equation}
        AoR = \frac{2h}{d_{base}}
        \label{eq:aor}
\end{equation}

We define the optimisation problem as follows: let the simulation parameters $\theta$ for simulator $p$ be such that $A_{sim}(\theta) = S(\theta)$, which is the simulated task measurement for static powder flowability in our case. 
$A_{real}$ is the target measurement from the real system ($\mathcal{M}_{real}$).
Let $\mathcal{H} = \{(\theta_i, E_i)\}$ be the history of evaluated parameters and their errors.
The objective is then to solve Equation~\ref{eq:sim2realgap}.
To achieve this, we use Bayesian Optimisation (BO)~\cite{Snoek2012}, as illustrated in Algorithm~\ref{alg:optimisation}.
The process begins by initialising ($\mathcal{I}_B$) the search bounds for $\theta$ based on simulator constraints (Table~\ref{table:parameters}).
At each iteration $k$, new parameters $\theta_{k+1}$ ($\mathcal{O}$) are proposed, $E_{k+1}$ is evaluated through $S(\theta_{k+1})$, and $\mathcal{H}$ is updated.
If progress stagnates after $T_{stagnation}$ iterations, the search bounds are adaptively refined ($\mathcal{B}_h$) using the range of the top 10\% best-performing parameter sets in $\mathcal{H}$.
The process continues until termination conditions $\mathcal{E}_S$ are met, aiming to collect $N$ acceptable parameter sets in $\Theta_{opt}$.

% \begin{algorithm}
%     \caption{Flowability optimisation loop}
%     \label{alg:optimisation}
%     \begin{algorithmic}[1]
%         \Require $N \geq 1$
%         \State $options \gets \emptyset$
%         \State $results \gets \emptyset$
%         \State $A_{real} = REAL\_LIFE\_MEASUREMENT()$
%         \State $i \gets 0$
%         \State $counter \gets 0$
%         \State $INITIALISE\_BOUNDS()$
%         \State $\theta \gets \emptyset$
%         \State $E \gets \infty$
%         \While{$N \geq 1$}
%             \State $\theta \gets SUGGEST\_PARAMETERS(\theta, E)$
%             \State $A_{sim} \gets SIMULATE(\theta)$
%             \State $E \gets |A_{real} - A_{sim}|$
%             \State $results.add(\{\theta, E\})$
%             \If{$E<1$}
%                 \State $options.add(\theta)$
%                 \State $N \gets N-1$
%             \ElsIf{$counter \geq 100$}
%                 \State $REDUCE\_BOUNDS(results)$
%                 \If{$EXHAUSTED\_SEARCHSPACE()$}
%                     \State \Return $options$
%                 \EndIf
%                 \State $counter \gets 0$
%             \EndIf
%             \State $counter \gets counter+1$
%         \EndWhile
%         \State \Return $options$
%     \end{algorithmic}
% \end{algorithm}

% Optional: Define symbols used

\begin{algorithm}[htbp]
    \caption{\small {Sim-to-real Optimisation Using Powder Flowability}}
    \label{alg:optimisation}
    \begin{algorithmic}[1]
        \Require $N \geq 1$ (Number of solutions desired)
        \State $\Theta_{opt} \gets \emptyset$
        \State $\mathcal{H} \gets \emptyset$
        \State $A_{real} \gets \mathcal{M}_{real}()$
        % \Comment{Obtain real measurement}
        \State $k \gets 0$
        \State $\mathcal{I}_B()$ 
        % \Comment{Initialize search bounds}
        \State $\theta \gets \emptyset$ 
        % \Comment{Initial parameter set (may be determined by $\mathcal{I}_B$ or $\mathcal{P}$)}
        \State $E \gets \infty$ 
        % \Comment{Initial error}
        \While{$N > 0$}
            \State $\theta \gets \mathcal{O}(\theta, E)$ 
            % \Comment{Suggest new parameters}
            \State $A_{sim} \gets p(\theta)$ 
            % \Comment{Run simulation}
            \State $E \gets |A_{real} - A_{sim}|$ 
            % \Comment{Calculate error}
            \State $\mathcal{H} \gets \mathcal{H} \cup \{(\theta, E)\}$ 
            % \Comment{Store result}
            \If{$E < 1$} 
            % \Comment{Check if error is below threshold}
                \State $\Theta_{opt} \gets \Theta_{opt} \cup \{\theta\}$ 
                % \Comment{Add parameters to solution set}
                \State $N \gets N - 1$ 
                % \Comment{Decrement count of solutions needed}
            \ElsIf{$k \geq T_{stagnation}$} 
            % \Comment{Check if bound reduction interval reached}
                \State $\mathcal{B}_{h}(\mathcal{H})$ 
                % \Comment{Reduce search bounds}
                \If{$\mathcal{E}_S()$} 
                % \Comment{Check if search space exhausted}
                    \State \Return $\Theta_{opt}$ 
                    % \Comment{Return found solutions if search exhausted}
                \EndIf
                \State $k \gets 0$ 
                % \Comment{Reset interval counter}
            \EndIf
            \State $k \gets k + 1$ 
            % \Comment{Increment iteration counter}
        \EndWhile
        \State \Return $\Theta_{opt}$ 
        % \Comment{Return found solutions}
    \end{algorithmic}
\end{algorithm}
\begin{table}[h!]
    \caption{Physical parameter bounds for optimisation of powder simulation in NVIDIA Isaac Sim.}
    \centering
    \resizebox{0.8\linewidth}{!}{%
        \begin{NiceTabular}{*{3}{c}}[hlines]
            \CodeBefore
                % \rowcolors[gray]{1}{0.9}{1}[respect-blocks]
                % \rowcolor{NavyBlue!50}{1}
            \Body
            \textbf{Physical Parameter} & \textbf{Minimum} & \textbf{Maximum} \\ 
            Particle Diameter [mm] & 0.22 & 0.32 \\
            Adhesion [$\cdot$] & 0.0 & 1.3 \\
            Particle Adhesion Scale [$\cdot$] & 0.0 & 1.3\\
            Adhesion Offset Scale [$\cdot$] & 0.0 & 0.00005\\
            Friction [$\cdot$] & 0.0 & 1.3\\
            Gravity Scale [$\cdot$] & 0.3 & 1.1 \\
            Damping [$\cdot$] & 0.0 & 1.0 \\
            Cohesion [$\cdot$] & 0.0 & 1.2 \\
            Particle Mass [$\mu$g] & 13 & 16.5
        \end{NiceTabular}
    }
    
    \label{table:parameters}
\end{table}

\subsubsection{Simulation Assumptions}
% Describe the simplifiucations we do within the simulator in order to mimic the the powders     

Due to the complexity of particle physics, certain assumptions and simplifications were adopted in our simulation. 
First, particles are assumed to be perfectly spherical.
While real powders exhibit complex geometries, this simplification accelerates collision detection computations, making the simulation practical for efficient training of robotic manipulation policies.
Second, we assumed that all particles within a given simulation instance possess identical, uniform size and mass, thus neglecting the inherent polydispersity found in most real-world powders. 
The particle diameter is constrained to lie within a range that is computationally tractable. 
Consequently, the permissible particle mass range is determined based on these dimensions and the densities of the specific materials being simulated.
Third, particle dynamics are simulated using the Position-Based Dynamics (PBD) solver~\cite{Mueller2007}.
Although this framework simulates the dynamics of both inter-particle and external forces applied to the system, it satisfies constraints such as attachments and contacts using a position-level approach that prioritises simulation speed at the expense of physical fidelity.
The physical parameters governing these interactions, listed in Table~\ref{table:parameters}, are bounded by physically plausible real-world values and the inherent limits of the simulator (NVIDIA Isaac Sim~\cite{NVIDIA2025}).
It should be noted that due to the simplifying assumptions employed (\textit{e.g.}, spherical shape, uniform size, PBD approximations), the optimal values identified for these parameters may not directly correspond to their real-world physical counterparts.
Our objective is not to estimate exact micro-dynamic coefficients but rather to calibrate the simulation parameters to accurately reproduce the macroscopic powder behaviour.

\subsection{Flowability-Informed Powder Weighing}
\label{ssec:FLIP_method}

\subsubsection{Problem Formulation}

We use the notion of a Markov Decision Process (MDP)~\cite{barto} to formulate the powder weighing task as a reinforcement learning problem. 
The MDP is represented as a tuple $ \mathcal{M} = \langle \mathcal{X, A, P, R ,}  \gamma \rangle$ of a set of states ($\mathcal{X}$), actions ($\mathcal{A}$), transition probabilities ($\mathcal{P}$), and immediate rewards ($\mathcal{R}$), with $\gamma$ being a scalar discount factor in $[0,1]$, that determines the emphasis given to immediate rewards relative to future ones. 
We use parametrised shake and incline robot motions, similar to the method in~\cite{powder_weighing_tanaka}, to improve sample efficiency and accelerate policy convergence.
We define the observation $o$ as the triplet $o=(w_{current},w_{target}, \theta_{spoon})$, $w_{target}$ represents the target amount of powder, $w_{current}$ the quantity of powder that has been dumped so far, $\theta_{spoon}$ the current pitch of the spoon. 
The action $\mathcal{A}$ space is continuous and two-dimensional, $a = (a_{shake}, a_{incline} \in \mathcal{A})$.
Here, $a_{shake}$ specifies the amplitude of the shaking motion (backward displacement before returning to the initial position) and $a_{incline}$ dictates the target inclination angle (pitch) for the spoon tilt. 
The reward function is computed from the error between the target weight $w_{target}$ and the dispensed weight $w_{current}$ calculated as follows:

\begin{equation}
    \mathcal{R} = \begin{cases}
             \Delta +1, & \Delta > -1 \\  
             \Delta,  & otherwise
        \end{cases},
    \label{eq:reward}
\end{equation}
where $\Delta=-|w_{target}-w_{current}|$.

\subsubsection{Flowability-Optimised Data} \label{sssection:random_method}

The static and dynamic flow characteristics of a powder impact the precision achievable during robotic weighing. 
Materials with high flowability allow more controlled dispensing, whereas cohesive powders are prone to adhesion, clumping, and erratic bulk discharge, which increases the task complexity. 
Since flowability is an emergent property—that is, different combinations of micro-parameters like friction, cohesion, and particle geometry can yield similar flow behaviour—it serves as a crucial indicator of manipulation difficulty. 
Therefore, we structure our agent training around this macroscopic property. 
Instead of broadly randomising individual simulation particle parameters, we focus the training distribution by sampling particle parameters conditioned on achieving a diverse range of target flowability values. 
This flowability-conditioned training enhances data efficiency by grouping physically distinct powders that present similar control challenges. 
We hypothesise that policies trained under this regime will generalise effectively to novel powders based on their flowability, rather than requiring precise knowledge of the underlying micro-parameters.

To this aim, let $F_{r}:[AoR_{min}, AoR_{max}]$ be the representative flowability range over which we wish to train our policy. 
We assume that for each flowability level $F$ sampled from this range, there exists a physics dynamics model represented by the state transition function $p_F(s_t|s_{t-1},a_{t-1})$. 
Within a parameterised simulation, this function becomes $p(s_t|s_{t-1}, a{t-1}, \theta_F)$, for which  the parameter set $\theta_F$ can be derived by the method in Section~\ref{sssec:optimisation_process}. 
Constructing a flowability-informed dataset then simply involves finding a set:
\begin{equation}
 \overline{\theta}_\mathcal{F}=\{\theta_{F_1}, \theta_{F_2} ,\theta_{F_3}, ... , \theta_{F_{N-1}}, \theta_{F_N}\}
\end{equation}
where $\mathcal{F}=\{F_1, F_2, F_3, ..., F_N\}$ represents the sampled flowability levels from our range $R$, which could be obtained through random uniform sampling or curriculum sampling (Section~\ref{sssection:curriculum_method}).
% Algorithm \ref{alg:random} summarises the training process using random uniform sampling of the optimised data. 
% \input{algorithms/random_training}

\subsubsection{Flowability-Informed Curriculum Learning} \label{sssection:curriculum_method}

We employ a curriculum learning strategy~\cite{Bengio2009} to accelerate policy training within simulation environments configured using flowability-optimised parameters (derived via Algorithm~\ref{alg:optimisation}).
The curriculum is structured based on powder flowability, ordered from highest flowability (corresponding to the lowest AoR) to lowest flowability (highest AoR).
Formally, this corresponds to starting from simulation parameters $\theta \in \Theta^*_{\mathcal{F}}$ associated with the highest flowability regime within our target set $\mathcal{F}$.
The curriculum then progressively transitions to parameter sets representing lower flowability levels, thereby systematically increasing the task difficulty for the learning agent.
This easy-to-hard progression is motivated by the empirical observation that low-flowability powders, often exhibiting significant cohesion and adhesion, present greater challenges for precise robotic manipulation and mass control during the weighing task.

As detailed in Algorithm~\ref{alg:curriculum_symb}, this curriculum is implemented by first indexing the optimised parameter sets $\Theta^*_{\mathcal{F}}$ by their associated flowability metric (\textit{e.g.}, AoR).
At each episode, the training environment is configured with parameters ($\theta$) representing a powder of a specific flowability level. 
After each episode, the error ($e_i$) ,between the target amount and the amount dumped, is evaluated. 
The curriculum advances to the subsequent level (lower flowability) only when the policy achieves a predetermined performance criterion, assessed by the error over the last K episodes, at the present level (line 22).
To mitigate potential stagnation on a particularly challenging flowability class, a timeout mechanism is incorporated: the curriculum advances to the subsequent parameter set if the performance criteria are not met within a predefined maximum of $M$ episodes for the current level, regardless of the achieved error (line 27).
% We define the policy performance as satisfactory when the final error in the last 20 episodes is below 1mg for all episodes, or when their mean error is below 0.8mg. 
\vspace{-0.5em}
% \begin{algorithm}
%     \caption{Flowability-Informed Powder Weighing}
%     \label{alg:curriculum}
%     \begin{algorithmic}[1]
%         \Require
%             \Statex \textit{Sampled flowability levels} $\mathcal{F}$
%             \Statex \textit{Configurations per sample} $N$
%         \Statex  
%         \State Collect $\overline{\theta_\mathcal{F}}$ using Algorithm \ref{alg:optimisation}
%         \State $SORT(\mathcal{F})$
%         \State $level \gets 0$
%         \State \textit{Initialise policy} $\pi$
%         \State \textit{Initialise FIFO queue} $errors$
%         \For{$i \gets 0$ \textit{to MAX EPISODE NUMBER}}
%             \State $\theta_F \gets \overline{\theta_\mathcal{F}}[level]$
%             \State \textit{Set simulation parameters to} $\theta_F$
%             \For{$j \gets 0 $} \textit{to MAX EPISODE STEP}
%                 \State \textit{Get action} $a_t$ \textit{from} $\pi$
%                 \State $STEP\_ENVIRONMENT(a_t$)
%                 \State \textit{Add transition to replay buffer}
%                 \State \textit{Update} $\pi$         
%             \EndFor
%             \State \textit{Compute episode error} $e$
%             \If{$len(errors)==20$}
%                 \State $errors.pop()$
%             \EndIf
%             \State $errors.push(e)$
%             \If{$mean(errors)<0.8$ OR $all(errors)<1$}
%                 \State $level \gets level+1$
%                 \State $errors.empty()$
%             \EndIf
%         \EndFor
        
%     \end{algorithmic}-
% \end{algorithm}

\begin{algorithm}[htbp]
    \caption{Flowability-informed Powder Weighing \textbf{\textit{(FLIP)}}}
    \label{alg:curriculum_symb}
    \begin{algorithmic}[1]
        \Require Flowability indexed parameter sets  $\Theta^*_{\mathcal{F}} = \{\theta^{(0)}, \dots, \theta^{(L-1)}\}$; Max episodes $N_{max}$, Max steps $T_{max}$; Error thresholds $T_{mean}$, $T_{max\_err}$; Error queue size $K$, Max episodes per level $M$
        % Assumes L levels, sorted easy-to-hard

        \State $\pi \leftarrow \pi_0$
        % \Statex \Comment{Initialize policy}
        \State $\mathcal{D} \leftarrow \emptyset$
        % \Statex \Comment{Initialize replay buffer}
        \State $\mathcal{Q}_E \leftarrow \text{FIFOQueue}(K)$
        % \Statex \Comment{Initialize error queue of size K}
        \State $level \leftarrow 0$
        % \Statex \Comment{Initialize curriculum level index}
        \State $m \leftarrow 0$
        % \Statex \Comment{Initialize episode counter for current level}

        \For{$i \gets 1$ to $N_{max}$}
            \State $\theta \leftarrow \text{SampleConfig}(\Theta^*_{\mathcal{F}},level)$
            % \Statex \Comment{Get parameters for current level}
            \State $\text{SetConfig}(\theta)$
            % \Statex \Comment{Configure simulator}
            \State $s \leftarrow \text{EnvReset}()$
            % \Statex \Comment{Reset environment, get initial state $s_0$}
            \For{$t \gets 1$ to $T_{max}$}
                \State $a \leftarrow \pi(s)$
                % \Statex \Comment{Select action}
                \State $s', r, d \leftarrow \text{EnvStep}(a)$
                % \Statex \Comment{Execute action}
                \State $\mathcal{D} \leftarrow \mathcal{D} \cup \{(s, a, r, s', d)\}$
                % \Statex \Comment{Store transition}
                \State $\pi \leftarrow \text{UpdatePolicy}(\mathcal{D})$
                % \Statex \Comment{Update policy}
                \State $s \leftarrow s'$
                % \Statex \Comment{Update state}
                % \If{$d$}
                %     \State \textbf{break}
                % \EndIf
            \EndFor

            \State $e_i \leftarrow \text{GetEpisodeError}(s)$ % Assumes error derived from final state/trajectory
            % \Statex \Comment{Compute final error for episode i}
            \State $\mathcal{Q}_E.\text{push}(e_i)$
            % \Statex \Comment{Add error to queue (handles FIFO logic implicitly)}

            \State $m \leftarrow m + 1$
            % \Statex \Comment{Increment episode counter for current level}

            \If{$|\mathcal{Q}_E| = K$} 
                \If{$\text{Mean}(\mathcal{Q}_E) < T_{mean}$ \textbf{or} $\text{Max}(\mathcal{Q}_E) < T_{max\_err}$}
                % \Statex \Comment{Check if performance criteria met (queue full AND condition met)}
                    \State $level \leftarrow \min(level + 1, L-1)$ % Advance level, prevent overflow
                    % \Statex \Comment{Advance to next curriculum level}
                    \State $\mathcal{Q}_E.\text{clear}()$
                    % \Statex \Comment{Clear error queue for new level}
                    \State $m \leftarrow 0$
                    % \Statex \Comment{Reset episode counter for new level}
                \EndIf
            \ElsIf{$m \ge M$}
            % \Statex \Comment{Check if stagnation timeout reached}
                 \State $level \leftarrow \min(level + 1, L-1)$ % Advance level regardless
                 % \Statex \Comment{Advance to next curriculum level (timeout)}
                 \State $\mathcal{Q}_E.\text{clear}()$
                 % \Statex \Comment{Clear error queue}
                 \State $m \leftarrow 0$
                 % \Statex \Comment{Reset episode counter}
            \EndIf

            % \If{$level = L-1$ \textbf{and} $|\mathcal{Q}_E| = K$ \textbf{and} ($\text{Mean}(\mathcal{Q}_E) < T_{mean}$ \textbf{or} $\text{Max}(\mathcal{Q}_E) < T_{max\_err}$)}
            %  % \Statex \Comment{Optional: Check if final level criteria met to potentially break early}
            %  \State \textbf{break} % Exit outer loop if final level is mastered
            % \EndIf

        \EndFor
        \State \Return $\pi$
        % \Statex \Comment{Return trained policy}
    \end{algorithmic}
\end{algorithm}

\section{Experimental Evaluation}
\label{section:exp_evaluation}
We evaluated our flowability-informed powder weighing method both in simulation and in real world experiments designed to answer the following questions:
(1) Can an automated system accurately collect powder flowability data comparable to manual methods?
(2) Can real-world flowability data be integrated into simulation to model diverse powders accurately?
(3) Does our~\textbf{\textit{FLIP}} method improve robot policy learning for autonomous powder weighing in simulation and reality?
Our initial experiments validated simulation fidelity by comparing simulated powder dynamics to empirical flowability data, essential for successful sim-to-real transfer. 
Subsequent experiments compared \textbf{\textit{FLIP}}-trained robot policies (using different data ordering strategies) against those trained without flowability information, evaluating task completion and weighing accuracy.
\vspace{-0.7em}
\subsection{Experimental Setup}
% \subsubsection{Experimental Platform}
All of our simulation environments were developed in IsaacLab~\cite{mittal2023orbit} and are outlined below (Section~\ref{sssection:autmatic-static_flow} and~\ref{sssection:robotic-powder_weighing}).
These were run on a system featuring a 13th Gen Intel(R) Core(TM) i7-13700F processor, 64GB of RAM and an NVIDIA GeForce RTX 4090 GPU, operating on Ubuntu 20.04. 
A Franka Research 3 (FR3) robotic manipulator in all simulation and real-world setups.

\subsubsection{Autonomous Robotic Flowability Measurement}
\label{sssection:autmatic-static_flow}

We automated the static powder flowability measurement for reliable execution in both real-world and simulation environments (Fig.~\ref{fig:flip_pipeline}). 
The robot grasps and inverts the powder platform, pouring the contents into a closed stainless steel funnel with a $\SI {14} {mm}$ exit diameter.
It then centres the platform beneath the funnel.
Powder release is controlled by an Arduino Uno-actuated servo-motor, which opens a slide gate at the funnel's bottom.
An RGB-D camera (\textit{Intel RealSense D405}) is used to measure the height of the pile that forms on the platform. 
The RGB image is processed to locate the funnel's tip ($p_{tip}$) and the powder pile's apex ($p_{apex}$) (Fig.~\ref{fig:static_flowability}), and their 3D coordinates are extracted using the corresponding depth data. 
The height of the pile ($h$) is computed by subtracting the measured distance between these two points from the known distance between the funnel and the platform ($D$) as detailed in Equation \ref{eq:powder height}.
\begin{equation}
    \label{eq:powder height}
    h = D - \| \vec{p}_{\text{apex}} - \vec{p}_{\text{tip}} \|
\end{equation}
We use Equation~\ref{eq:aor} to compute the AoR, where the diameter of the pile is the diameter of its base. 
% Our robotic flowability measurement system using Isaac Orbit~\cite{mittal2023orbit} as shown in Fig.~\ref{fig:flip_pipeline}. 
In the simulated environment the height of the pile can directly be computed using the 3D coordinates of each independent particle. 

% \begin{figure}[H]
%     \centering
%     \includegraphics[width=0.8\linewidth]{images/Salt Working Example_croped.png}
%     \caption{The red point indicates the 3D coordinate of the funnel tip, while the blue point shows the 3D coordinate of the powder pile apex. The distance between these two points (here, 43.5mm) is measured and subtracted from the known platform-to-funnel distance (60mm).}
%     \label{fig:automatic_calculation}
% \end{figure}

\subsubsection{Robotic Powder Weighing}
\label{sssection:robotic-powder_weighing}
% Show it with the vial adapter or with the paper cone? 
As illustrated in Fig.~\ref{fig:flip_pipeline}, our experimental setup incorporates a FR3 robot arm equipped with a Robotiq 85F Gripper.
The gripper is mounted parallel to the working surface to maneouver the dispensing spoon more easily. 
The spoon operates above a 20ml vial, which serves as the target container. 
To gauge the powder amount in the vial, a precision scale (Fisherbrand FPRS223) is used, feeding the measured weight directly into the \textbf{\textit{FLIP}} framework.   
% A paper cone is positioned around the vial to ensure that any overflow lands outside the scale's weighing area, thus enabling precise measurement of the contents within the target vessel.
While the real-world setup is largely replicated in simulation, known simulator constraints with the vial's intricate CAD mesh led to collision issues. 
Consequently, the model was simplified to a comparable-sized cuboid for improved collision detection and computational efficiency.
% Second, the paper cone was excluded, as the simulator allows for direct computation of the powder's weight in the vial.

% \vspace{-2.1em}
\subsection{Powder Flowability Optimisation Results}
\label{ssection:flowabilit-optimisation}
To evaluate the performance of our robotic system for automatic AoR measurement (Section~\ref{sssection:autmatic-static_flow}), a comparative study was conducted against manual measurements performed according to the ISO 8398:1989 standard. 
We tested a range of powders with AoRs (\textit{i.e.,} different flowabilities) spanning $27^\circ$ to $43^\circ$. 
For each powder sample, five manual measurements and five automated system measurements were recorded.
The differences between the measured AoR values obtained by each method are reported in Table~\ref{table:aor_errors}. 
The largest discrepancy between manual and automated measurements was observed for semolina, with an absolute error of $1.8^\circ$. 
Across all powders, the average absolute error was $0.84^\circ$, demonstrating the reliability and repeatability of our automated measurement process. 
% These results support its use as an efficient and consistent alternative to manual methods.

% \begin{figure}[H]
%     \centering
%     \includegraphics[width=0.99\linewidth]{images/aor_errors.png}
%     \caption{A comparison of manual and automated AoR static powder flowability measurement.}
%     \label{fig:aor_errors}
% \end{figure}

\begin{table}[htbp] 
    \caption{Comparison of manual and automated AoR static powder flowability measurements. }
    \label{table:aor_errors} 
    \centering
    \resizebox{0.95\linewidth}{!}{% % Resizes to 95% of the column width
        \begin{NiceTabular}{l c c}[hlines] % Use NiceTabular with hlines option
            \CodeBefore % Code to execute before the body
                % \rowcolor{NavyBlue!50}{1} % Apply color to the first row (header)
            \Body % Start of the table body
            \textbf{Material} & \textbf{Automated AoR ($^\circ$)} & \textbf{Manual AoR ($^\circ$)} \\
            Sand & 28.55 $\pm$ 1.94 & 27.69 $\pm$ 2.54 \\
            Sugar & 31.41 $\pm$ 1.42 & 30.93 $\pm$ 0.87 \\
            Salt & 36.03 $\pm$ 2.29 & 35.8 $\pm$ 1.71 \\
            \hline
            Semolina & 42.762 $\pm$ 1.97 & 40.91 $\pm$ 1.85 \\
            Sodium Bicarbonate & 42.38 $\pm$ 2.48 & 43.14 $\pm$ 1.0 \\
        \end{NiceTabular}
    }
\end{table}

These automated AoR values are then used as target flowability levels in a simulation parameter optimisation process. 
For each powder, we perform a BO search to identify simulation parameters that yield a simulated AoR within a threshold of $T_{\text{AoR}} = 1.5^\circ$ from the measured value. 
We retain the top 10 parameter sets that satisfy this constraint.
For each powder, Table~\ref{table:parameters} presents visual comparisons of the simulated and real pile formations, and includes the mean and standard deviation of the absolute AoR error derived from the 10 highest-ranked solutions.
% , and the total number of evaluations required to discover them. 

\begin{table*}[h!]
    \caption{Real-world data and the resulting optimised powder simulation. The sim-to-real error measured for the five different powder samples demonstrates the fidelity achieved by parameter optimisation.}
    \centering
    \resizebox{0.9\linewidth}{!}{%
        \begin{NiceTabular}{*{6}{c}}[hlines]
            & \textbf{Sand} & \textbf{Sugar} & \textbf{Salt} & \textbf{Semolina} & \textbf{Sodium Bicarbonate} \\ 
            Real powder pile & \includegraphics[width=0.15\textwidth]{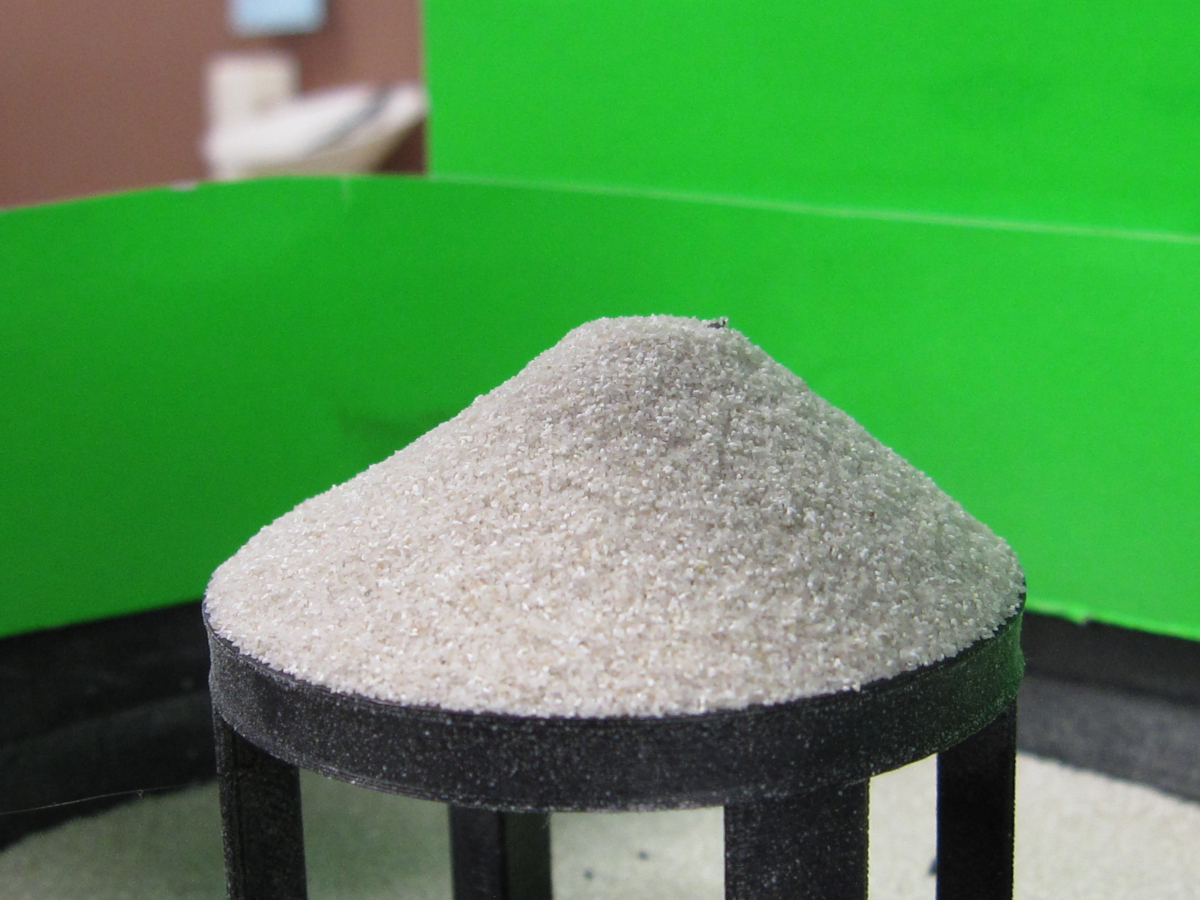} & \includegraphics[width=0.15\textwidth]{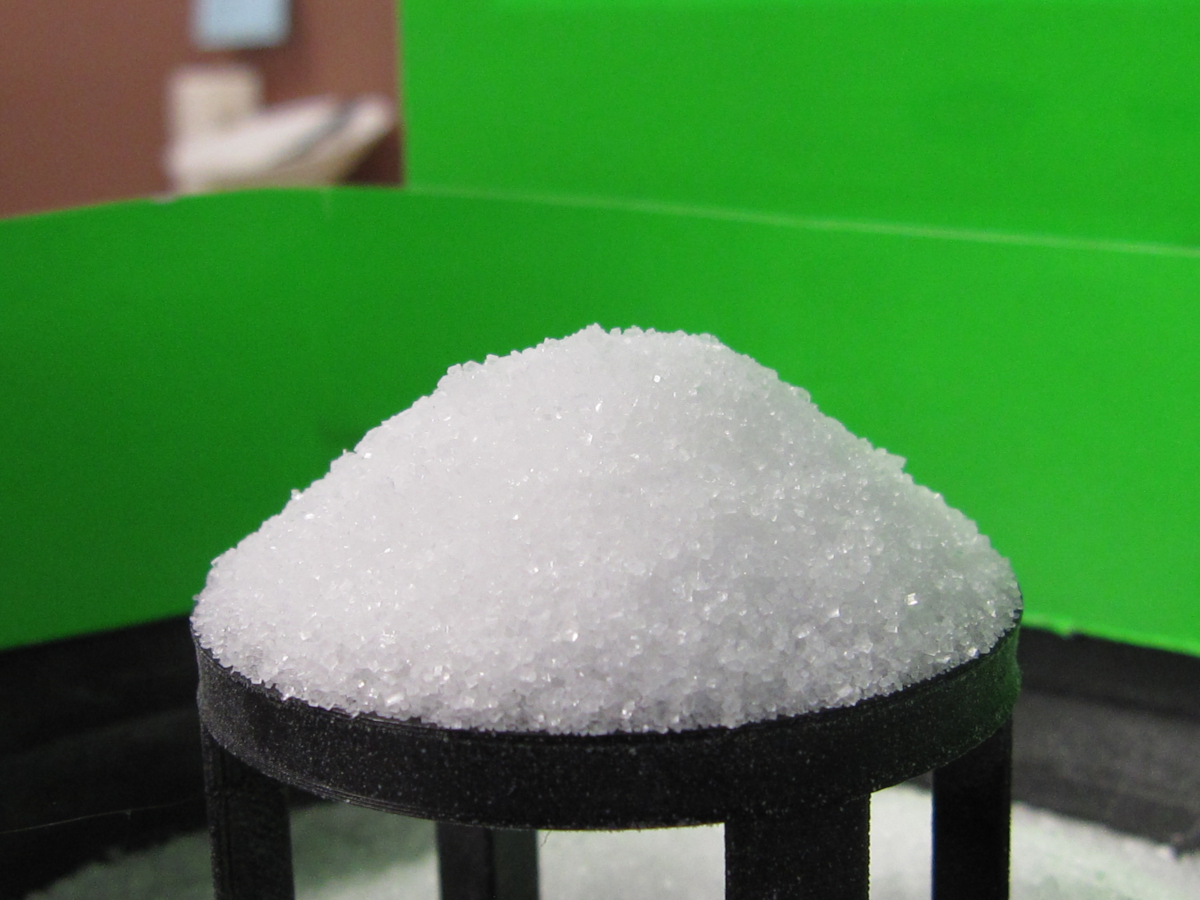}  & \includegraphics[width=0.15\textwidth]{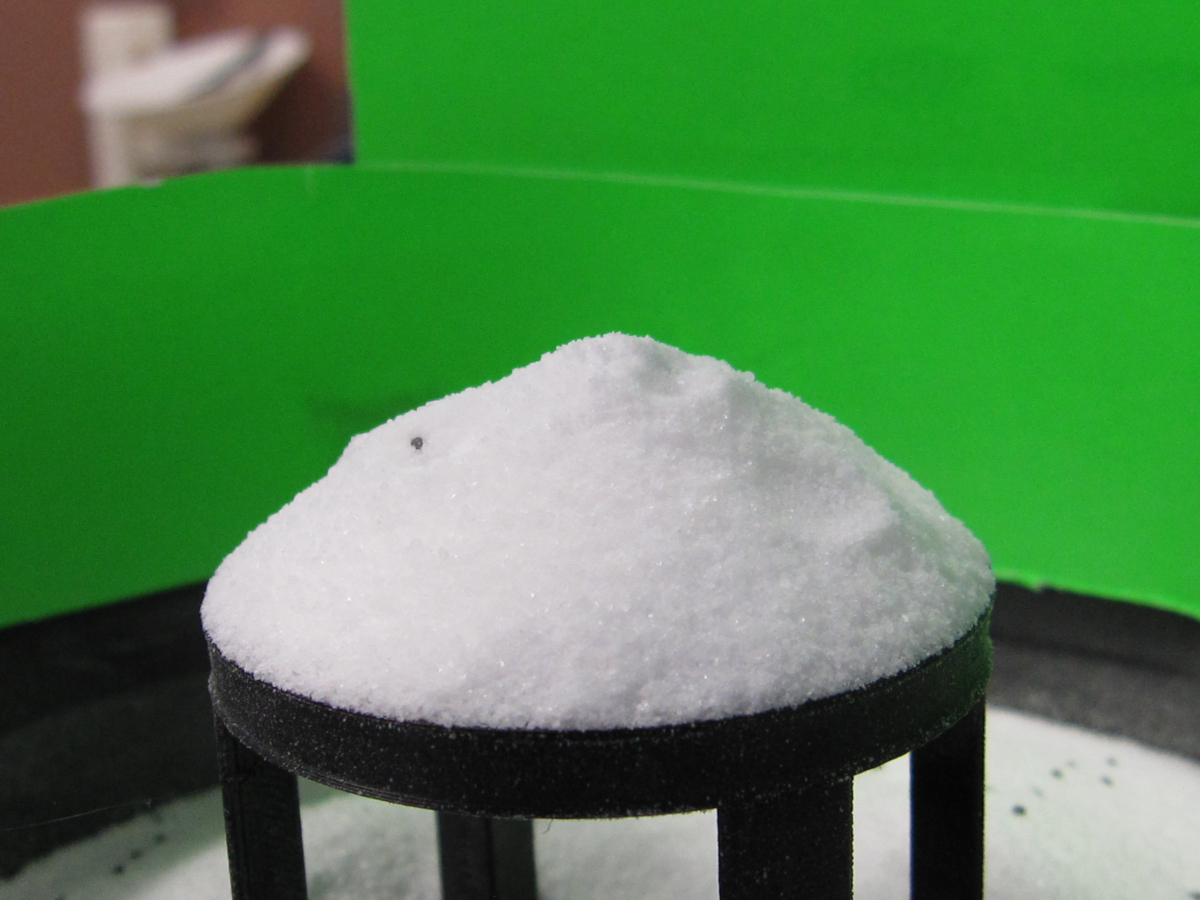} & \includegraphics[width=0.15\textwidth]{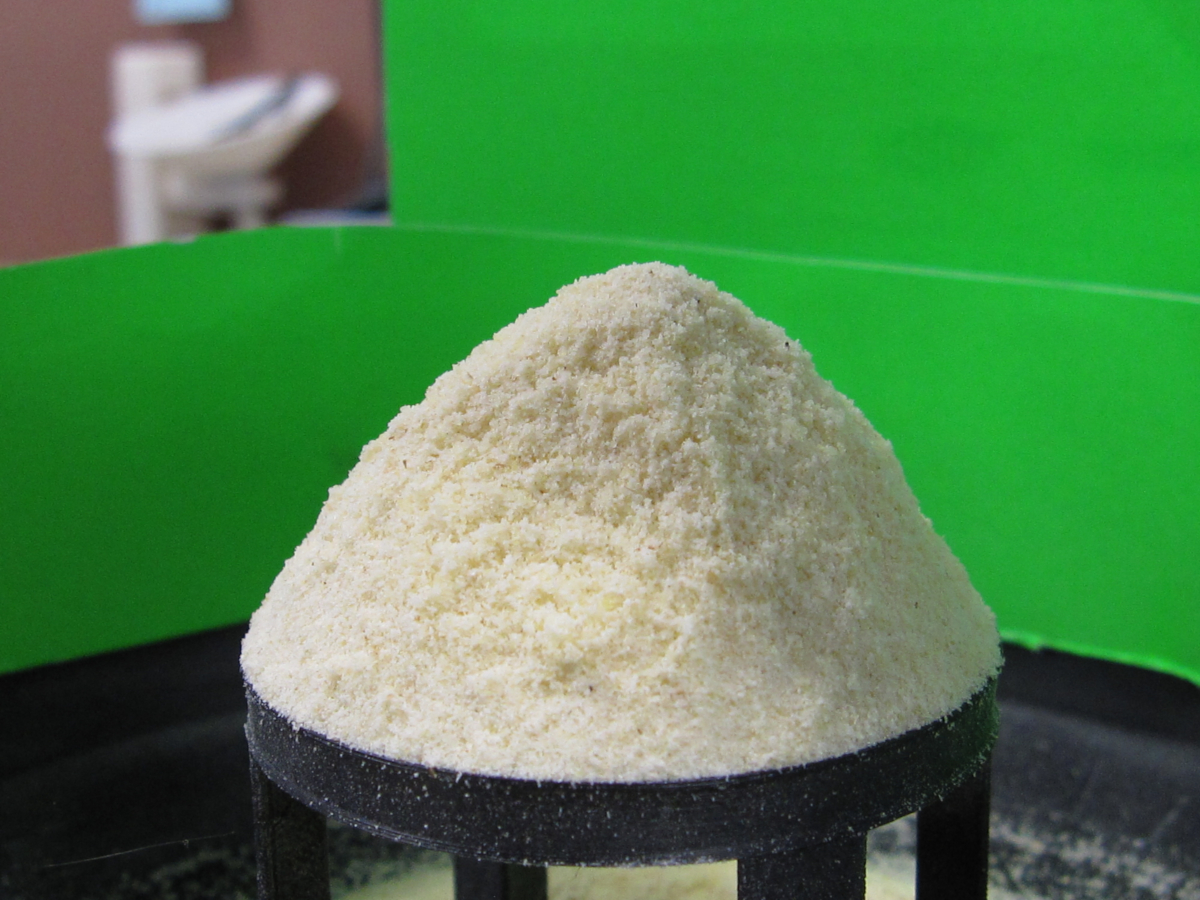} & \includegraphics[width=0.15\textwidth]{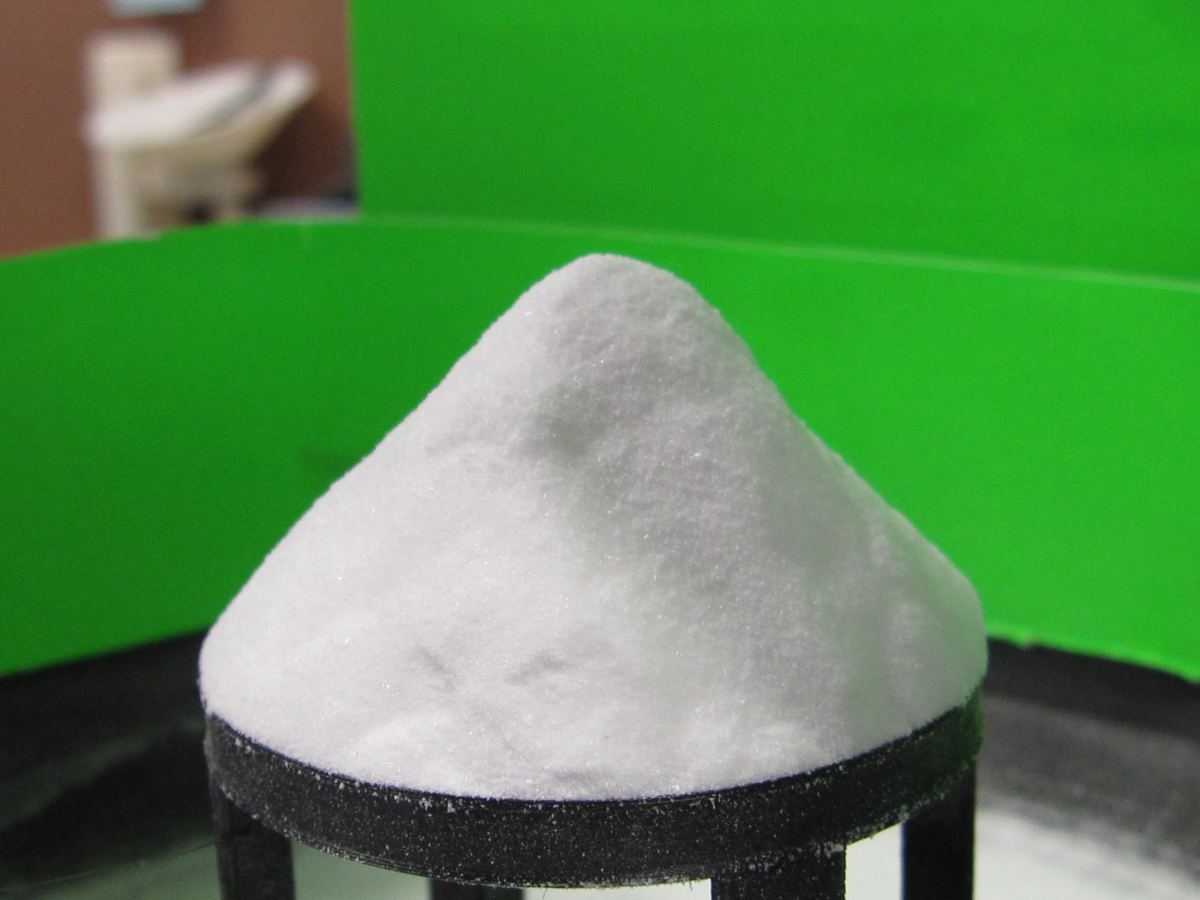}\\
            Simulated powder pile &\includegraphics[width=0.15\textwidth]{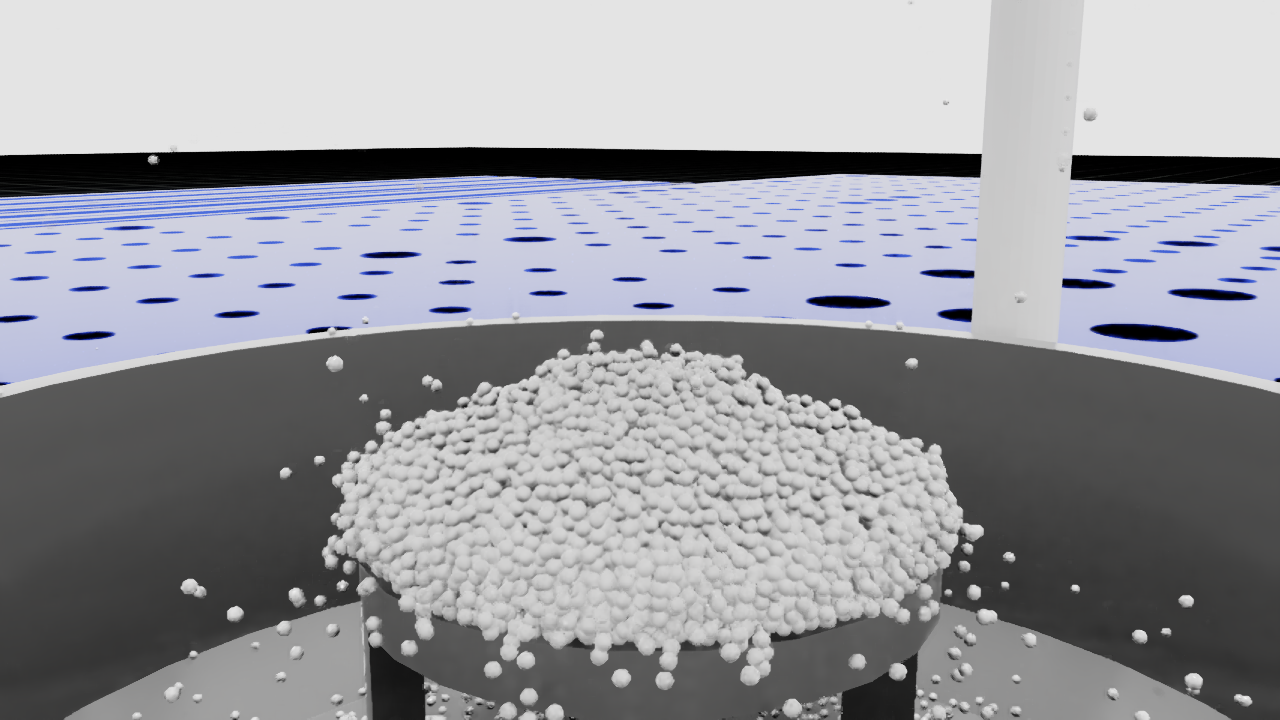} & \includegraphics[width=0.15\textwidth]{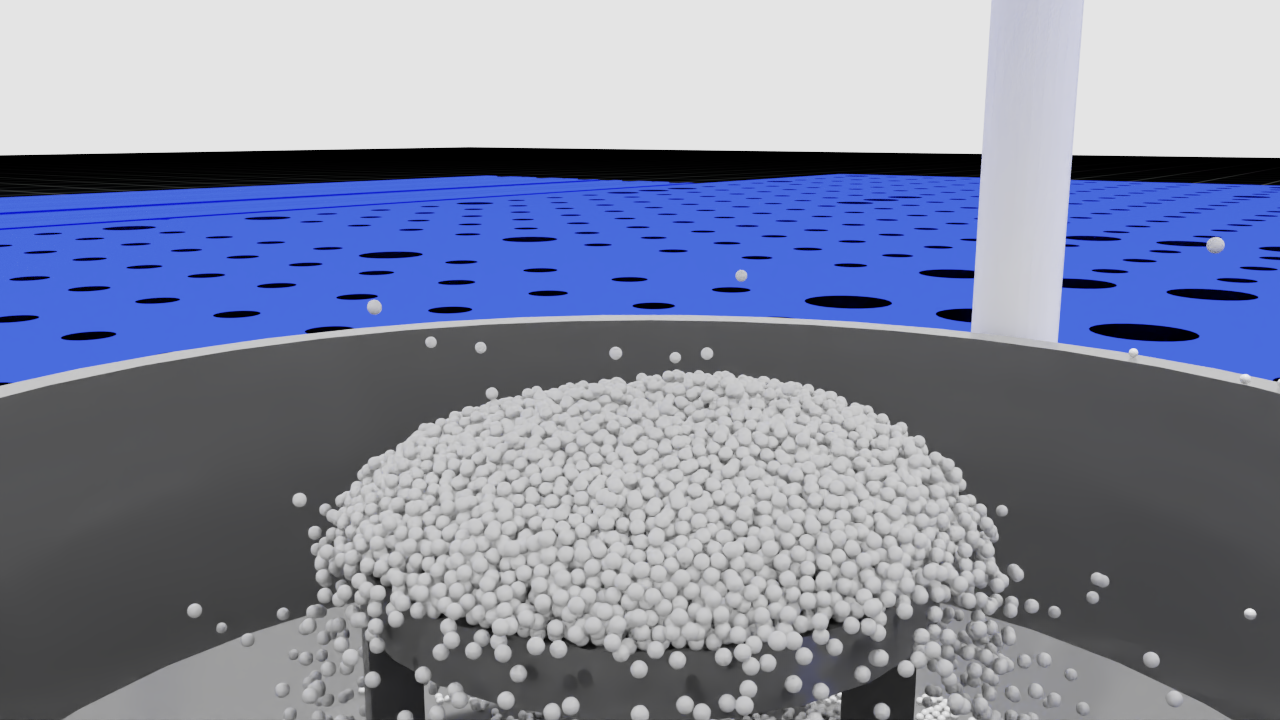}  & \includegraphics[width=0.15\textwidth]{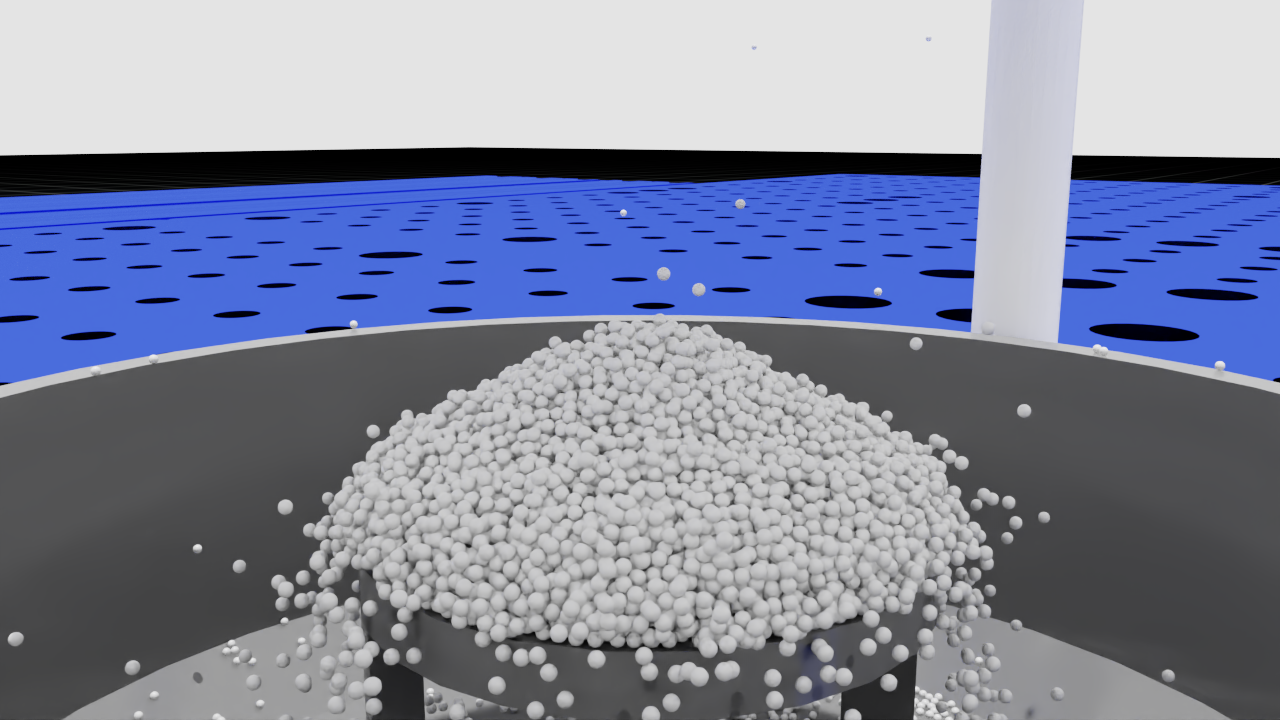} & \includegraphics[width=0.15\textwidth]{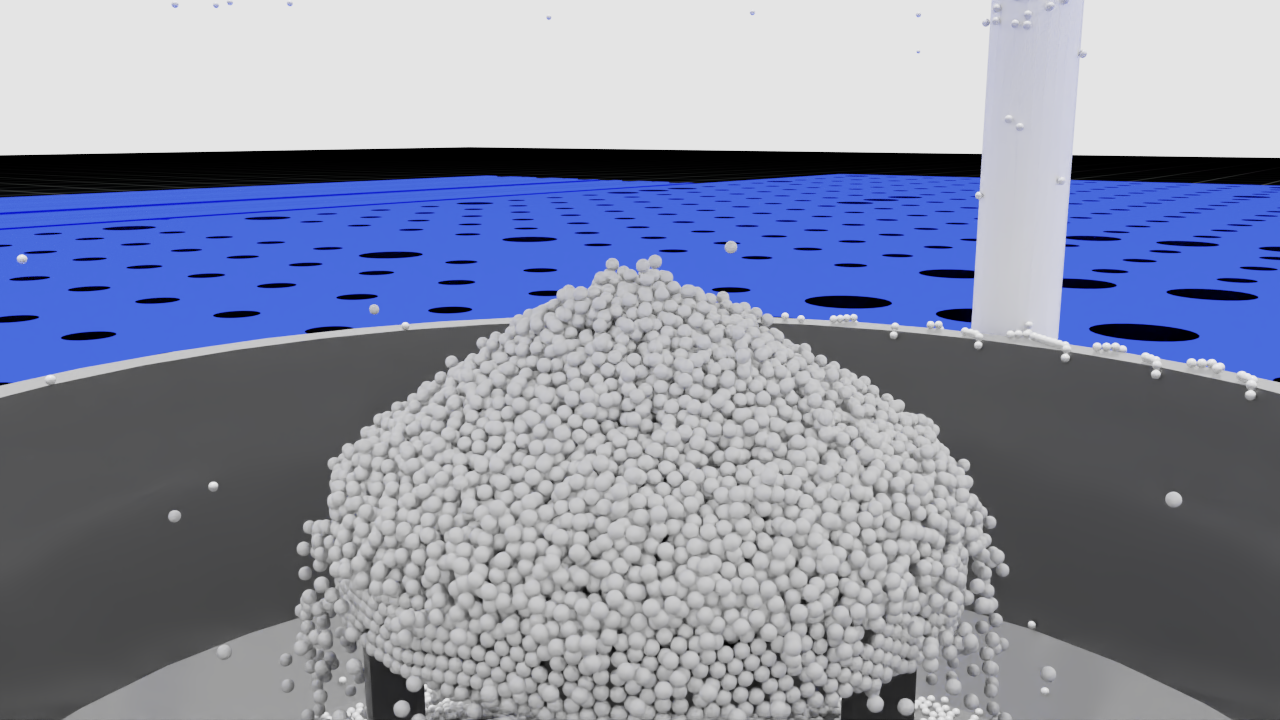} & \includegraphics[width=0.15\textwidth]{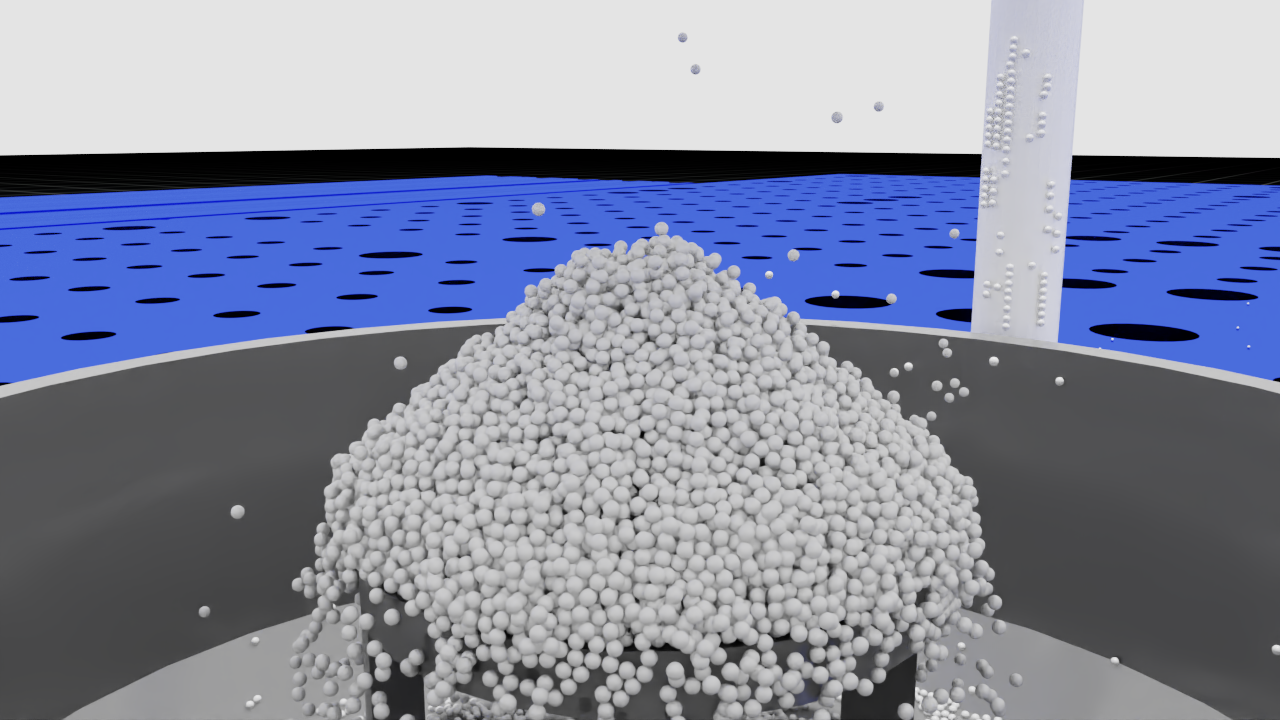}\\
            Sim-to-real error & $0.25\pm 0.13$  & $0.22\pm 0.12$ &$1.09\pm0.15$ & $0.40 \pm 0.14$ &$1.37 \pm0.43$ \\
            % No. Evaluations & 52 &  64 & 137 &144 & 150
        \end{NiceTabular}
 }   
    \label{table:parameters}
\end{table*}

\begin{table*}[htbp!]
    \caption{Real-world performance comparison for the powder weighing task. For each powder sample, results per method represent the mean absolute error and the corresponding standard deviation, calculated over 5 runs using the best-performing policy. Sodium bicarbonate and semolina represent out-of-distribution materials; the policy was not trained on these specific powders or materials with similar flow characteristics. The $\SI{20}\mg$ target weight was also untrained.}
    \centering
    \resizebox{0.95\linewidth}{!}{%
        \begin{NiceTabular}{l *{10}{c}}[hlines, cell-space-limits=3pt]
            \hline % Using a single hline at the very top
            \Block{2-1}{\textbf{Method}} & \multicolumn{10}{c}{\textbf{Powder Weighing Error (mg)}} \\
            % Removed the cmidrule here
             & \multicolumn{2}{c}{Salt} & \multicolumn{2}{c}{Sugar} & \multicolumn{2}{c}{Sand} & \multicolumn{2}{c}{Sodium Bicarbonate} & \multicolumn{2}{c}{Semolina}\\
            \hline % A single hline to separate the material/weight headers from the data
             & \SI{15}\mg & \SI{20}\mg & \SI{15}\mg & \SI{20}\mg & \SI{15}\mg & \SI{20}\mg & \SI{15}\mg & \SI{20}\mg & \SI{15}\mg & \SI{20}\mg \\
            \midrule % Keeping midrule for separation before the data rows
            Domain Randomisation & $3.2 \pm 1.93$ & $16.5\pm5.1$ & $2.83\pm1.3$ & $4.0\pm 3.0$ & $2.33\pm 1.4$ & $1.83\pm0.6$ & $7.8\pm4.83$ & $7.0\pm6.89$ & $6.4\pm4.17$ & $9.2\pm4.79$\\
            \textbf{\textit{FLIP}} (Random)& $2.6\pm2.33$ & $4.8\pm3.05$ & $2.0\pm1.09$ & $2.66\pm1.24$ & $\mathbf{1.33\pm0.47}$ & $\mathbf{1.81\pm1.83}$ & $9.4\pm6.52$ & $4.8\pm3.31$& $4.0\pm1.78$ & $6.0\pm3.79$\\
            \textbf{\textit{FLIP}} (Curriculum)& $\mathbf{2.4\pm1.86}$ & $\mathbf{2.4\pm1.74}$ & $\mathbf{1.6\pm0.48}$ & $\mathbf{1.2\pm0.97}$ & $1.6\pm 0.8$ & $2.2\pm1.32$ & $\mathbf{3.6\pm2.3}$ & $\mathbf{1.6\pm1.35}$& $\mathbf{1.8\pm0.97}$ & $\mathbf{2.8\pm2.31}$\\
            \textbf{\textit{FLIP}} (Reverse Curriculum)& $13.6\pm7.9$ & $11.0\pm8.87$ & $11.4\pm5.35$ & $1.8\pm1.16$ & $3.8 \pm 2.22$ & $5.8\pm3.13$ & $12.2\pm3.34$ & $8.6\pm6.71$& $9.5\pm5.52$ & $11.2\pm7.65$\\
            \bottomrule % Keeping bottomrule at the very end
        \end{NiceTabular}
    }
    \label{table:sim2real_results}
\end{table*}

\subsection{Simulation Powder Weighing  Results}
\label{ssection:simulation-experiment}

We evaluate our \textbf{\textit{FLIP}} method using three different orderings of flowability-optimised data: random ordering, curriculum ordering (\textit{i.e.,} most to least flowable) and reverse curriculum ordering (\textit{i.e.,} least to most flowable).
These are compared against a baseline method employing domain randomisation across all physical parameters, comparable to previous works~\cite{powder_weighing_tanaka}.
We define the training flowability range as $F_r \in [28^\circ, 37^\circ]$ and sample three flowability levels: $28^\circ$, $33^\circ$, and $37^\circ$, corresponding to sand, sugar, and salt.
This range purposely excludes highly cohesive, flour-like materials such as semolina and sodium bicarbonate.
% and 
% In simulation, we train both random and curriculum-trained agents using the flowability-optimised dataset described in Section~\ref{sssection:random_method}.
% Additionally, we train two baseline policies: a domain randomisation policy and a reverse curriculum policy. The Domain Randomization policy samples parameters across the entire supported range of the physics engine (Table~\ref{table:parameters}). The Reverse Curriculum policy follows a similar approach to Section~\ref{sssection:curriculum_method}, but progresses from the least to the most flowable powders.  
We use a model-free, off-policy DRL algorithm, Soft Actor-Critic (SAC)~\cite{Haarnoja2018} to train our agent.
We adopted the same neural network architecture and implementation, including a training termination criterion of 4000 episodes, as in related works~\cite{powder_weighing_tanaka} for a fair comparison.
For curriculum ordering, training for each flowability level had a cut-off of 1330 episodes.
A transition to the next level occurred earlier if the mean final error over the last 20 episodes fell below \SI{0.8}{\mg} or if the maximum final error fell below \SI{1}\mg.
We treated these thresholds as hyperparameters and manually tuned them for the task. 
Other hyperparameters included the number of optimised data points per flowability level (set to 7, a value found to balance learning quality and stability), and  the choice of flowability levels themselves (based on real-world data availability).
% , a learning rate of $1\times10^{-4}$, and a batch size of 16.
Powder weighing error (in \SI{}\mg) is the primary evaluation metric, averaged over 5 independent runs with different random seeds. 
\begin{figure}[h!]
    \begin{subfigure}[a]{\columnwidth}
        \includegraphics[width=0.92\linewidth]{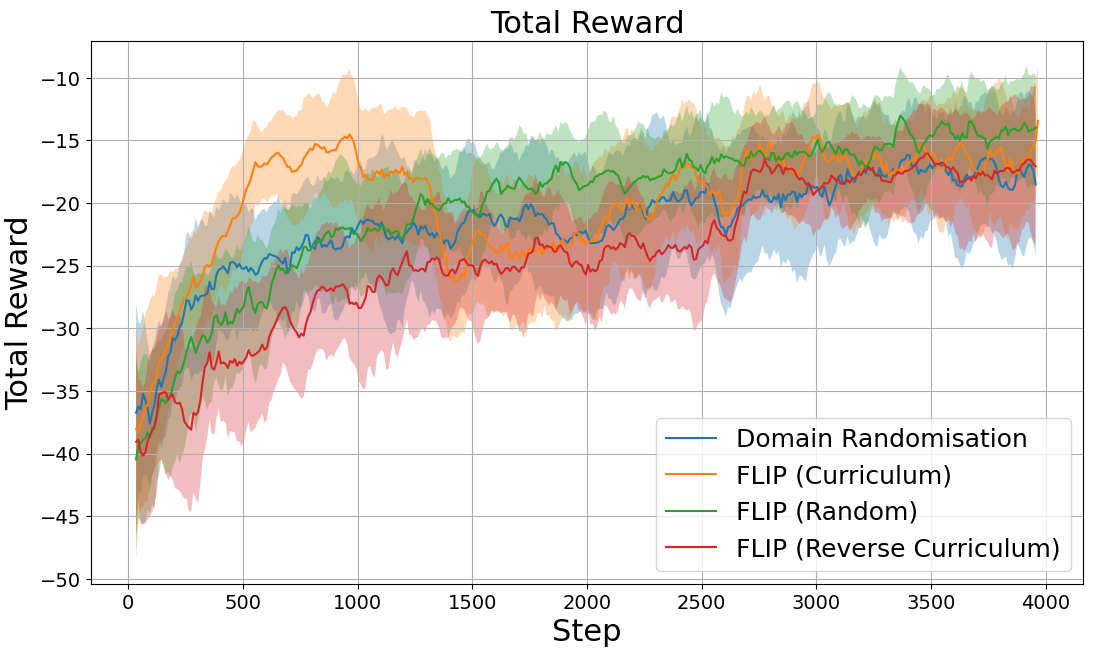}
        \caption{}
        \label{fig:total_reward}
    \end{subfigure}
    \begin{subfigure}[a]{\columnwidth}
        \includegraphics[width=0.92\linewidth]{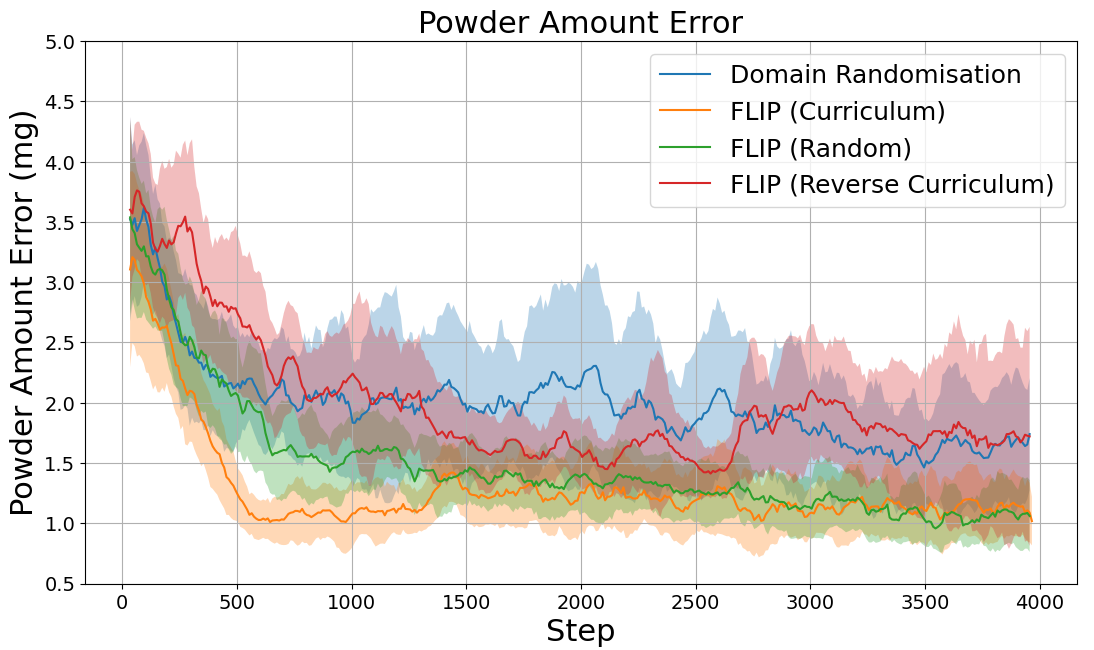}
        \caption{}
        \label{fig:final_error}
    \end{subfigure}
    \caption{Total reward (a) and final powder weighing error (b), averaged over 5 random seeds.}
    \label{figure:graphs}
\end{figure}
% Each run takes approximately 4.6 days.

Our results are presented in Fig.~\ref{figure:graphs}. 
\textbf{\textit{FLIP}} with random and curriculum data ordering methods are the only ones achieving close to $\SI {1} \mg$ accuracy, with the curriculum-based method converging faster.
It is worth noting that the curriculum agent reaches mg-level accuracy after approximately 600 episodes at the initial flowability level. 
The total reward curve (Fig.~\ref{fig:total_reward}) shows a dip when transitioning from $28^\circ$ to $33^\circ$, although the final error remains below $\SI {1.5}{\mg}$. 
This likely results from increased cohesion at higher flowability levels, which reduces powder displacement per action, requiring more steps to dispense the desired amount. 
Conversely, the reverse curriculum exhibits a decrease in total error when switching to highly flowable powders, but an increase in final error due to overshooting. 
The domain randomisation method shows high instability, evidenced by a large standard deviation in final powder weighing errors.

\subsection{Real-world Powder Weighing Results}
We conduct sim-to-real experiments to evaluate the performance of our method \textbf{\textit{FLIP}} in the context of an autonomous powder weighing task. 
For each evaluation, we select the policy corresponding to the best-performing run in simulation, where performance is measured by the lowest final error over the evaluation window. 
The trained network is transferred zero-shot to the real robotic system, with the robot speed and shake amplitude calibrated to match those used in the simulator. 
Each selected policy is tested across a set of five powders (two of which have out-of-distribution flowability levels) with varying flowabilities (see Table~\ref{table:parameters}) and for two target weights: \SI{15}\mg\ and \SI{20}\mg;  the latter target weight was not seen previously during training. 
For each configuration, we repeat the task five times and report the mean and standard deviation of the absolute error. 
% The powders used in this evaluation are the same as those described in Section~\ref{}.
% The objective of this experiment is to assess the zero-shot transfer capabilities of the trained policies and demonstrate their robustness to real-world conditions associated with autonomous powder handling.
Results are presented in Table~\ref{table:sim2real_results}, where the best-performing policy for each condition is highlighted. 
Our method, \textbf{\textit{FLIP}}, outperforms both the baseline method and the ablation using reverse curriculum policy across all tested materials.
The curriculum \textbf{\textit{FLIP}} policy achieves the lowest average powder weighing error across all trials ($2.12 \pm \SI{1.53}\mg$), followed by the random \textbf{\textit{FLIP}} policy ($3.94 \pm 2.55$), the DR policy without flowability data ($6.11 \pm 3.92$) and finally, the reverse curriculum \textbf{\textit{FLIP}} policy ($8.19 \pm 4.88$).
The curriculum-based policy demonstrates strong generalisation to previously unseen, more cohesive materials (out-of-distribution flowability levels).
For example, sodium bicarbonate, with an AoR of $42^\circ$, lies outside the training range of $28^\circ$ to $37^\circ$, yet is handled effectively. 
Given that the reverse curriculum policy performs the worst overall, even worse than no flowability data as is the DR method, our results demonstrate that using flowability data such that an agent is trained to face progressively more cohesive materials is beneficial. 
This underscores the critical importance of both accurate simulation calibration and an appropriate learning strategy in realising robust robotic manipulation policies for powder handling.

% for all but one material, across both target weights. 

% While training was performed only on target weights between \SI{5}\mg\ and \SI{15}\mg, the policy still achieves less than \SI{2.}\mg\ error for all powders at the \SI{20}\mg\ target weight. 
% Moreover, it performs well on . 

% This robustness may stem from a known discrepancy in the simulator: due to continuous particle motion even after the material has settled, the measured AoR tends to be underestimated.
% This can lead to a potential overestimation of simulated powder flowability, which appears more pronounced in certain cases. 
% As such, simulated powders may behave similarly to real powders with a higher AoR than their measured simulated value suggests.

\section{Conclusions}
This paper presented a new approach to robotic powder weighing in autonomous laboratories through the first successful application of flowability-informed reinforcement learning. 
Our core contribution is \textbf{\textit{FLIP}}, a method that strategically integrates powder flowability information into both simulation design and curriculum learning. 
This is supported by a successful simulation-to-reality transfer pipeline that uses real-world flowability metrics, including automated AoR measurements, to create high-fidelity simulation training environments for improved robotic manipulation of powders.
Both simulation and real-world experiments demonstrated that policies trained with \textit{\textbf{FLIP}} outperform baselines lacking flowability information, achieving low powder weighing errors and strong generalisation even to more cohesive, unseen materials and target weights. 
This work represents a step forward in sample-efficient and robust sim-to-real transfer for robotic handling of granular materials, paving the way for more reliable autonomous laboratory systems. 
Future work will explore how the current robot controller impacts task performance, particularly focusing on novel adaptations to better accommodate cohesive powders ($AoR > 45$), where current displacements are insufficient. 
Ultimately, extending this approach to complementary tasks like powder scooping tasks could establish a comprehensive and robust end-to-end autonomous powder weighing pipeline but also fundamentally transform how novel, uncharacterised materials are handled in chemical research labs. More generally, this approach might have analogous uses in other lab automation domains where approximate physical models are available.
% Additionally, we plan to refine the curriculum learning mechanism by considering total reward progression—as it may provide a more informative switching signal than error alone in certain plateaus.

% add future work mention? 
\vspace{-0.7em}
\section*{Acknowledgements}
This work was supported by the Leverhulme Trust through the Leverhulme Research Centre for Functional Materials Design, the Royal Academy of Engineering under the
Research Fellowship Scheme and EPSRC through the AI for chemistry: AIchemy hub (EPSRC grant EP/Y028775/1 and EP/Y028759/1).
A.I.C. thanks the Royal Society for a Research Professorship (RSRP\textbackslash S2\textbackslash 232003).
All authors would like to thank Dr. Ben Alston and Dr. Charlotte Boott for their input on the materials chemistry.

%%%%%%%%%%%%%%%%%%%%%%%%%%%%%%%%%%%%%%%%%%%%%%%%%%%%%%%%%%%%%%%%%%%%%%%%%%%%%%%%

\bibliographystyle{IEEEtran}
\bibliography{bibliography}

\begin{thebibliography}{10}
\providecommand{\url}[1]{#1}
\csname url@samestyle\endcsname
\providecommand{\newblock}{\relax}
\providecommand{\bibinfo}[2]{#2}
\providecommand{\BIBentrySTDinterwordspacing}{\spaceskip=0pt\relax}
\providecommand{\BIBentryALTinterwordstretchfactor}{4}
\providecommand{\BIBentryALTinterwordspacing}{\spaceskip=\fontdimen2\font plus
\BIBentryALTinterwordstretchfactor\fontdimen3\font minus \fontdimen4\font\relax}
\providecommand{\BIBforeignlanguage}[2]{{%
\expandafter\ifx\csname l@#1\endcsname\relax
\typeout{** WARNING: IEEEtran.bst: No hyphenation pattern has been}%
\typeout{** loaded for the language `#1'. Using the pattern for}%
\typeout{** the default language instead.}%
\else
\language=\csname l@#1\endcsname
\fi
#2}}
\providecommand{\BIBdecl}{\relax}
\BIBdecl

\bibitem{Tom2024}
G.~Tom, S.~P. Schmid, S.~G. Baird, Y.~Cao, K.~Darvish, H.~Hao, S.~Lo, S.~Pablo-García, E.~M. Rajaonson, M.~Skreta, and et~al., ``Self-driving laboratories for chemistry and materials science,'' \emph{Chemical Reviews}, 2024.

\bibitem{Burger2020AMR}
\BIBentryALTinterwordspacing
B.~Burger, P.~M. Maffettone, V.~V. Gusev, C.~M. Aitchison, Y.~Bai, X.~yan Wang, X.~Li, B.~M. Alston, B.~Li, R.~Clowes, N.~Rankin, B.~Harris, R.~S. Sprick, and A.~I. Cooper, ``A mobile robotic chemist,'' \emph{Nature}, vol. 583, pp. 237 -- 241, 2020. [Online]. Available: \url{https://api.semanticscholar.org/CorpusID:220420261}
\BIBentrySTDinterwordspacing

\bibitem{Lunt2024}
\BIBentryALTinterwordspacing
A.~M. Lunt, H.~Fakhruldeen, G.~Pizzuto, L.~Longley, A.~White, N.~Rankin, R.~Clowes, B.~Alston, L.~Gigli, G.~M. Day, A.~I. Cooper, and S.~Y. Chong, ``Modular{,} multi-robot integration of laboratories: an autonomous workflow for solid-state chemistry,'' \emph{Chem. Sci.}, vol.~15, pp. 2456--2463, 2024. [Online]. Available: \url{http://dx.doi.org/10.1039/D3SC06206F}
\BIBentrySTDinterwordspacing

\bibitem{Dai2024}
\BIBentryALTinterwordspacing
T.~Dai, S.~Vijayakrishnan, F.~T. Szczypi{\'{n}}ski, J.-F. Ayme, E.~Simaei, T.~Fellowes, R.~Clowes, L.~Kotopanov, C.~E. Shields, Z.~Zhou, J.~W. Ward, and A.~I. Cooper, ``Autonomous mobile robots for exploratory synthetic chemistry,'' \emph{Nature}, Nov 2024. [Online]. Available: \url{https://doi.org/10.1038/s41586-024-08173-7}
\BIBentrySTDinterwordspacing

\bibitem{jiang}
\BIBentryALTinterwordspacing
Y.~Jiang, H.~Fakhruldeen, G.~Pizzuto, L.~Longley, A.~He, T.~Dai, R.~Clowes, N.~Rankin, and A.~I. Cooper, ``Autonomous biomimetic solid dispensing using a dual-arm robotic manipulator,'' \emph{Digital Discovery}, vol.~2, pp. 1733--1744, 2023. [Online]. Available: \url{http://dx.doi.org/10.1039/D3DD00075C}
\BIBentrySTDinterwordspacing

\bibitem{powder_weighing_tanaka}
Y.~Kadokawa, M.~Hamaya, and K.~Tanaka, ``Learning robotic powder weighing from simulation for laboratory automation,'' in \emph{2023 IEEE/RSJ International Conference on Intelligent Robots and Systems (IROS)}, 2023, pp. 2932--2939.

\bibitem{Prescott2000}
J.~Prescott and R.~Barnum, ``On powder flowability,'' \emph{Pharmaceutical Technology}, vol.~24, pp. 60--84+236, 01 2000.

\bibitem{Bengio2009}
\BIBentryALTinterwordspacing
Y.~Bengio, J.~Louradour, R.~Collobert, and J.~Weston, ``Curriculum learning,'' in \emph{Proceedings of the 26th Annual International Conference on Machine Learning}, ser. ICML '09.\hskip 1em plus 0.5em minus 0.4em\relax New York, NY, USA: Association for Computing Machinery, 2009, p. 41–48. [Online]. Available: \url{https://doi.org/10.1145/1553374.1553380}
\BIBentrySTDinterwordspacing

\bibitem{Zhou2025}
\BIBentryALTinterwordspacing
J.~Zhou, M.~Luo, L.~Chen, Q.~Zhu, S.~Jiang, F.~Zhang, W.~Shang, and J.~Jiang, ``A multi-robot–multi-task scheduling system for autonomous chemistry laboratories,'' \emph{Digital Discovery}, pp.~--, 2025. [Online]. Available: \url{http://dx.doi.org/10.1039/D4DD00313F}
\BIBentrySTDinterwordspacing

\bibitem{Darvish2025}
\BIBentryALTinterwordspacing
K.~Darvish, M.~Skreta, Y.~Zhao, N.~Yoshikawa, S.~Som, M.~Bogdanovic, Y.~Cao, H.~Hao, H.~Xu, A.~Aspuru-Guzik, A.~Garg, and F.~Shkurti, ``Organa: A robotic assistant for automated chemistry experimentation and characterization,'' \emph{Matter}, vol.~8, no.~2, p. 101897, 2025. [Online]. Available: \url{https://www.sciencedirect.com/science/article/pii/S2590238524005423}
\BIBentrySTDinterwordspacing

\bibitem{Butterworth2023}
A.~Butterworth, G.~Pizzuto, L.~Pecyna, A.~I. Cooper, and S.~Luo, ``Leveraging multi-modal sensing for robotic insertion tasks in r\&d laboratories,'' in \emph{2023 IEEE 19th International Conference on Automation Science and Engineering (CASE)}, 2023, pp. 1--8.

\bibitem{injection}
A.~Angelopoulos, M.~Verber, C.~McKinney, J.~Cahoon, and R.~Alterovitz, ``High-accuracy injection using a mobile manipulation robot for chemistry lab automation,'' in \emph{2023 IEEE/RSJ International Conference on Intelligent Robots and Systems (IROS)}, 2023, pp. 10\,102--10\,109.

\bibitem{Scamarcio2025}
\BIBentryALTinterwordspacing
V.~Scamarcio, J.~Tan, F.~Stellacci, and J.~Hughes, ``Reliable and robust robotic handling of microplates via computer vision and touch feedback,'' \emph{Frontiers in Robotics and AI}, vol.~11, 2025. [Online]. Available: \url{https://www.frontiersin.org/journals/robotics-and-ai/articles/10.3389/frobt.2024.1462717}
\BIBentrySTDinterwordspacing

\bibitem{pizzuto2022accelerating}
G.~Pizzuto, H.~Wang, H.~Fakhruldeen, B.~Peng, K.~S. Luck, and A.~I. Cooper, ``Accelerating laboratory automation through robot skill learning for sample scraping,'' 2022.

\bibitem{grinding}
Y.~Nakajima, M.~Hamaya, K.~Tanaka, T.~Hawai, F.~von Drigalski, Y.~Takeichi, Y.~Ushiku, and K.~Ono, ``Robotic powder grinding with audio-visual feedback for laboratory automation in materials science,'' 10 2023.

\bibitem{tobin2017domainrandomizationtransferringdeep}
\BIBentryALTinterwordspacing
J.~Tobin, R.~Fong, A.~Ray, J.~Schneider, W.~Zaremba, and P.~Abbeel, ``Domain randomization for transferring deep neural networks from simulation to the real world,'' 2017. [Online]. Available: \url{https://arxiv.org/abs/1703.06907}
\BIBentrySTDinterwordspacing

\bibitem{deformable_s2r}
P.~M. Scheikl, E.~Tagliabue, B.~Gyenes, M.~Wagner, D.~Dall'Alba, P.~Fiorini, and F.~Mathis-Ullrich, ``Sim-to-real transfer for visual reinforcement learning of deformable object manipulation for robot-assisted surgery,'' \emph{IEEE Robotics and Automation Letters}, vol.~8, no.~2, pp. 560--567, 2023.

\bibitem{Matl2020}
\BIBentryALTinterwordspacing
C.~Matl, Y.~S. Narang, R.~Bajcsy, F.~Ramos, and D.~Fox, ``Inferring the material properties of granular media for robotic tasks,'' \emph{2020 IEEE International Conference on Robotics and Automation (ICRA)}, pp. 2770--2777, 2020. [Online]. Available: \url{https://api.semanticscholar.org/CorpusID:212747555}
\BIBentrySTDinterwordspacing

\bibitem{stir-to-pour}
T.~Lopez-Guevara, R.~Pucci, N.~K. Taylor, M.~U. Gutmann, S.~Ramamoorthy, and K.~Suhr, ``Stir to pour: Efficient calibration of liquid properties for pouring actions,'' in \emph{2020 IEEE/RSJ International Conference on Intelligent Robots and Systems (IROS)}, 2020, pp. 5351--5357.

\bibitem{Liu2024}
\BIBentryALTinterwordspacing
W.~Liu, Z.~Deng, Y.~Zhang, X.~Zhu, J.~Huang, H.~Zhang, and J.~Zhu, ``Decoding powder flowability: Machine learning pioneers the analysis of particle-size distribution effects,'' \emph{Powder Technology}, vol. 435, p. 119407, 2024. [Online]. Available: \url{https://www.sciencedirect.com/science/article/pii/S0032591024000494}
\BIBentrySTDinterwordspacing

\bibitem{Snoek2012}
J.~Snoek, H.~Larochelle, and R.~P. Adams, ``Practical bayesian optimization of machine learning algorithms,'' in \emph{Proceedings of the 26th International Conference on Neural Information Processing Systems - Volume 2}, ser. NIPS'12.\hskip 1em plus 0.5em minus 0.4em\relax Red Hook, NY, USA: Curran Associates Inc., 2012, p. 2951–2959.

\bibitem{Mueller2007}
\BIBentryALTinterwordspacing
M.~M\"{u}ller, B.~Heidelberger, M.~Hennix, and J.~Ratcliff, ``Position based dynamics,'' \emph{J. Vis. Comun. Image Represent.}, vol.~18, no.~2, p. 109–118, Apr. 2007. [Online]. Available: \url{https://doi.org/10.1016/j.jvcir.2007.01.005}
\BIBentrySTDinterwordspacing

\bibitem{NVIDIA2025}
\BIBentryALTinterwordspacing
{NVIDIA}, ``{NVIDIA Isaac Sim},'' NVIDIA, 2025. [Online]. Available: \url{https://developer.nvidia.com/isaac-sim}
\BIBentrySTDinterwordspacing

\bibitem{barto}
R.~S. Sutton and A.~G. Barto, \emph{Reinforcement Learning: An Introduction}.\hskip 1em plus 0.5em minus 0.4em\relax Cambridge, MA, USA: A Bradford Book, 2018.

\bibitem{mittal2023orbit}
M.~Mittal, C.~Yu, Q.~Yu, J.~Liu, N.~Rudin, D.~Hoeller, J.~L. Yuan, P.~P. Tehrani, R.~Singh, Y.~Guo, H.~Mazhar, A.~Mandlekar, B.~Babich, G.~State, M.~Hutter, and A.~Garg, ``Orbit: A unified simulation framework for interactive robot learning environments,'' 2023.

\bibitem{Haarnoja2018}
T.~Haarnoja, A.~Zhou, P.~Abbeel, and S.~Levine, ``Soft actor-critic: Off-policy maximum entropy deep reinforcement learning with a stochastic actor,'' in \emph{ICML}, 2018.

\end{thebibliography}

\end{document}